\documentclass[a4paper,12pt]{article}

\usepackage[utf8]{inputenc}
\usepackage[english]{babel}
\usepackage{amsmath}
\usepackage{graphicx}
\usepackage[colorinlistoftodos]{todonotes}
\usepackage{float}
\usepackage{lipsum}
\usepackage{caption}

\usepackage{float}
\usepackage{mwe}
\usepackage{listings}
\usepackage{color}
\usepackage{physics}
\usepackage[english]{babel}
\usepackage{ragged2e}
\usepackage{blindtext}
\usepackage{afterpage}
\usepackage{ amssymb }
\usepackage{listings}

\usepackage{graphicx}
\usepackage{subcaption}
\usepackage{boxhandler}
\usepackage[section]{placeins}
\usepackage[a4paper, margin=2.5cm]{geometry}

\definecolor{dkgreen}{rgb}{0,0.6,0}
\definecolor{gray}{rgb}{0.5,0.5,0.5}
\definecolor{mauve}{rgb}{0.58,0,0.82}

\begin{document}

\pagenumbering{gobble}
\begin{figure}[ht]
   \minipage{0.76\textwidth}
		\includegraphics[width=4cm]{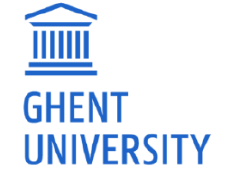}
		\label{EscudoUNAM}
   \endminipage
   \minipage{0.32\textwidth}
		\includegraphics[height = 3cm ,width=3.5cm]{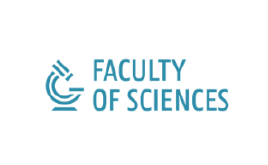}
		\label{EscudoFC}
	\endminipage
\end{figure}

\begin{center}
\vspace{3cm}

\fontseries{bx}
\fontsize{12pt}{12pt}
\selectfont
Using Machine Learning Methods for Automation of Size Grid Building and Management\\ 

\fontseries{bx}
\fontsize{12pt}{12pt}
\vspace{2cm}
\selectfont
 Salim Yunus\\

\end{center}

\begin{FlushRight}
\vspace{4cm}
Master dissertation submitted to\\
\vspace{0.25cm}
obtain the degree of\\
\vspace{0.25cm}
Master of Statistical Data Analysis\\

\vspace{0.75cm}
Promoter: Prof. Dries Benoit\\
\vspace{0.25cm}
Department of Marketing, Innovation and Organisation\\

\vspace{0.75cm}
Promoter: Filipa Peleja\\
\vspace{0.25cm}
Department of Data Analytics and AI\\

\vspace{0.75cm}
\textbf{Academic year : 2020-2021}

\end{FlushRight}

\clearpage\null\newpage

\begin{figure}[ht]
   \minipage{0.76\textwidth}
		\includegraphics[width=4cm]{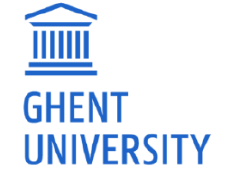}
   \endminipage
   \minipage{0.32\textwidth}
		\includegraphics[height = 3cm ,width=3.5cm]{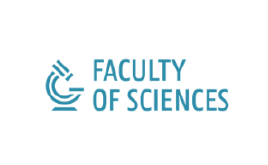}
	\endminipage
\end{figure}

\begin{center}
\vspace{3cm}

\fontseries{bx}
\fontsize{12pt}{12pt}
\selectfont
Using Machine Learning Methods for Automation of Size Grid Building and Management\\ 

\fontseries{bx}
\fontsize{12pt}{12pt}
\vspace{2cm}
\selectfont
 Salim Yunus\\

\end{center}

\begin{FlushRight}
\vspace{4cm}
Master dissertation submitted to\\
\vspace{0.25cm}
obtain the degree of\\
\vspace{0.25cm}
Master of Statistical Data Analysis\\

\vspace{0.75cm}
Promoter: Prof. Dries Benoit\\
\vspace{0.25cm}
Department of Marketing, Innovation and Organisation\\

\vspace{0.75cm}
Promoter: Filipa Peleja\\
\vspace{0.25cm}
Department of Data Analytics and AI\\

\vspace{0.75cm}
\textbf{Academic year : 2020-2021}

\end{FlushRight}

\newpage
The author and the promoter give permission to consult this master dissertation and to copy it or parts of it for personal use. Each other use falls under the restrictions of the copyright, in particular concerning the obligation to mention explicitly the source when using results of this master dissertation.

\newpage

\section*{Foreword}
This thesis was written for aiming completion to the master in Statistical Data Analysis at Ghent University. The programme is focusing on acquiring an advanced level of statistical knowledge and data analytical skills. The subject of this thesis falls within the scope of the programme's field since a data science approach was applied to a company's problem as part of its supply chain process. \\

The study was conducted in co-operation with Levi Strauss {\&} Co. Europe. The company is one of the market leaders in fashion and apparel industry, the operations is focusing on Europe market and was looking for an AI approach to one of its supply chain processes. The process is currently suffering from its workload for different teams and bottle neck problems in the processes where a more automated approach was demanded. Therefore, this study is aiming to build a model to automate the sizing process by using machine learning methods. A classification model was built to automate size grid building and management in the company. Satisfying results were obtained and the research will help the company on decision making for size selection for future season planning operations. The data used in the study is owned by the company and stored in the cloud system. The tables have millions of rows. Additionally, the data tables are confidential and are protected against unfair commercial use.\\

It has been 6 months that I have been conducting research on the topic. I have obtained more experience in supply chain in fashion and apparel industry and more importantly in machine learning. I would like to thank my supervisor from Ghent University, Prof. Dries Benoit and my supervisor from Levi Strauss, Filipa Peleja. Their valuable insights and directions gave me necessary guidance to complete the research. I am also thankful for my colleagues in the company who helped me learn more on the current structure and processes in the supply chain.\\

Salim Yunus\\

Ghent, August 15, 2021\\

\newpage

\tableofcontents

\cleardoublepage
\setcounter{page}{1}

\pagenumbering{arabic}

\newpage
\section*{Abstract}
Fashion apparel companies require planning for the next season, a year in advance for supply chain management. This study focuses on size selection decision making for Levi Strauss. Currently, the region and planning group level size grids are built and managed manually. The company suffers from the workload it creates for sizing, merchant and planning teams. This research is aiming to answer two research questions: "Which sizes should be available to the planners under each size grid name for the next season(s)?" and "Which sizes should be adopted for each planning group for the next season(s)?". We approach to the problem with a classification model, which is one of the popular models used in machine learning. With this research, a more automated process was created by using machine learning techniques. A decrease in workload of the teams in the company is expected after it is put into practice. Unlike many studies in the state of art for fashion and apparel industry, this study focuses on sizes where the stock keeping unit represents a product with a certain size.\\

\newpage

\addcontentsline{toc}{section}{1. INTRODUCTION}
\section*{1. INTRODUCTION}

\addcontentsline{toc}{section}{1.1 Motivation}
\subsection*{1.1 Motivation}
Fashion apparel industry is one of the fastest changing sectors. Fashion retailers are challenged by responding very quickly changing customer needs, foreseeing the new trends and fitting in the market as a competitor in such a speed industry. Despite all the fast dynamics in the sector causing the fluctuations in sales and making the product life cycles short, the lead times of products are long and can even be longer than a season which is defined for a product for how long it stays in the market. This forces fashion apparel companies to structure their supply chains very carefully and plan production in such a way that profitability is maximized, and environmental damage caused by disposal of unsold inventory is minimized [1]. Hence, demand forecasting models have become more significant for the fashion apparel companies. One of the components in demand forecasting for a clothing is decision making for which sizes to produce from a specific product for upcoming seasons. This study is focusing on the size selection decision making for a fashion apparel company called Levi Strauss{\&} Co.\

\addcontentsline{toc}{section}{1.2 Background}
\subsection*{1.2 Background}
In the company, the supply chain planning is at season level as Spring-Summer and Fall-Winter. One challenging task for the planning and sizing teams is to make decisions on selection of sizes for products based on the customer needs. The selection is not made on product level since it is difficult to manage ca. 2400-2500 different products in each season. Adding the number of sizes and customers that are geographically spread to European region, the supply chain structure becomes bigger and more complex. In order to overcome the complexity and management problems, products are grouped under the classes called as Size Grid Names. The decision making and the tools to maintain this process work on Size Grid Name level. In other words, Size Grid Name means a group of products with similar characteristics (fabric, consumer group, gender etc.). For each Size Grid Name, size grid, which is defined as the matrix showing all possible sizes of products that can be produced, is built for new products or maintained for existing products by the planners. Currently Size Grid Names are manually designed for each season, where the best size selection is tried to be adopted for each product. For this purpose, they collaborate with different teams to obtain qualitative and quantitative insights. Following this work, planners manually select the sizes for each size grid name where they try to make the optimal selection to ensure Levi’s consumers will have all the sizes demand fulfilled for the upcoming season. If a size grid is poorly designed, it will impact products not being sold or missing sell opportunities by not producing product/ size that corresponds to the actual consumer need, consequently, there will be profit loss.\

\addcontentsline{toc}{section}{1.3 Related Work}
\subsection*{1.3 Related Work}
Before size designation of clothes was invented in the early 1800’s, whether tailors or people themselves at home were making the clothes to fit on persons [2]. Size measurements are made based on anthropometric measurements. In time, different countries have developed different standards for sizes in clothes and even some countries have changed the size standards they developed in time. This has resulted in many different sizing standards in the world. However, in 21st century a lot of garments are being delivered by shipment. Because of the differences in size standards, many are being returned. This created a need to develop a sizing system in a more systematic way [3]. In order to create one standard, many studies are being conducted. European Standards Organization (CEN) produced some standards for size designation to create a national standard for 33 member states. In addition, The International Organization for Standardization has developed some standards and now they are being revised to be replaced with one of them which closely resembles one of European Standards. All the size designation standards are developed by measuring the human body dimensions. While sizes could be one dimension for some clothing products, some have two dimensions when one dimension is not enough [4]. The sizes of products in this study have either one dimension or two dimensions depending on the product category, which will be explained in the following sections.\\

Demand planning is one of the main challenges for the fashion retailers because of the short product life cycles, lack of history in data, very fluctuating customer demand, fast changing trends in fashion causing stock outs, high inventories, low service levels [5]. More machine learning techniques are being used to come up with a more robust model aiming to predict customer demand more accurately. The developed models are mainly focusing on product level, not on stock keeping unit (SKU) level. Stock keeping unit for the companies is a clothing product with certain fabric, design, style and more importantly with a size. Although some products can be reprocessed by cutting from bigger size to smaller ones, the other way around and for some products it is not possible to replenish them. In other words, SKU is produced, stored and purchased by the consumers. However, models are mostly developed in a way that SKU’s aggregated on product level. The reasons of developing demand planning models on product level rather than SKU could be the quantities on SKU level being too low, to have a more robust model, to get rid of some variance caused by trends. In [6], a demand forecasting model was developed where application of inventory constraints was addressed as a challenge. By aggregating SKU’s on product level and putting inventory and sales data on product level was the solution to overcome it.\

\addcontentsline{toc}{section}{1.4 Purpose and Research Questions}
\subsection*{1.4 Purpose and Research Questions}
Size grids are built and managed on size grid name level by the sizing team in the company. However, planners can select sizes on planning group level and size grid name level. Size grid building and management is critical to guarantee there’s no profit loss because of stock outs or excess inventory at the end of a season. The objective of this study is to automate size grid building and management by predicting the size of products that should be made available for upcoming seasons. In this regard, there are two research questions:\

\begin{itemize}
	\item Size grid name level: “Which sizes should be available under each size grid name for the next season(s)?”
    \item Planning group level: “Which sizes should be selected for each customer for the next season(s)?”
\end{itemize}

\addcontentsline{toc}{section}{1.5 Scope and Limitation}
\subsection*{1.5 Scope and Limitation}
This study has some limitations due to the structure of the company and its strategies:\

\begin{itemize}
	\item Although the company is operating globally, it has three different regions as America, Europe and Africa{\&} Asia. The study is conducted only for Europe region. In fact, results of this study can be a pilot and be expanded to other regions in the future.
    \item In the Europe region, there are many countries making the supply chain more complex and bigger. There are 10 affiliates, 4 distribution centers, 2 sale channels as retail and wholesale, 58 planning groups and 12,000 Sold-To’s (destination of products).
    \item Demand forecasting for products or SKU’s is not in the scope of this project as there is already an ongoing project for developing a demand planning model globally.
    \item A Size Grid Name is the representation of a group of products, where they were clustered manually by using some  attributes like gender, type, fit etc. There is not a scientific approach for segmenting them. In the scope of this project, size grid names will be accepted as they are. However, as a further research classification of products can be considered.
    \item There are two brands in the Europe region: Levi’s and Dockers. This study will be conducted only for Levi’s brand. It should also be noted that Dockers has a smaller share in terms of number of products sold and the revenue in the company.
    \item The company produces tops, bottoms, footwear and accessories. Footwear and accessories are out of the scope of this study. Size grids are currently built and managed only for tops and bottoms from product categories.
    \item In case a new size grid name is created, it won’t be in the scope of this study as there won’t be historical data for it. In fact, merchandising team is currently managing these cases by using insights from global and market data.
    \item In the history, size selections were made by the planners in the company. The decisions were used in the model as the target variable. This means there could have been more optimal size selections in the past. This study focuses on automation of the decisions made by the planners.
\end{itemize}

\addcontentsline{toc}{section}{1.6 Outline}
\subsection*{1.6 Outline}
In the next chapter, the process flow of size grid building and management will be explained. The required data tables for this study are presented with their characteristics and variables. Data cleaning and processing will be addressed afterwards. After analyzing the distributions of the variables and the relations between the variables, alternative model approaches will be discussed. In the following chapter, obtained results will be presented and discussed. Finally, an overview of the work will be summarized, and possible future work will be addressed.\\

\addcontentsline{toc}{section}{2. METHODS {\&} RESULTS}
\section*{2. METHODS {\&} RESULTS}
\subsection*{2.1 The current process of Size Grid Building and Management}

A size grid represents a matrix of sizes for a group of products. A group of products are classified based on product attributes such as gender and this group is called a size grid name. A size grid name has a string structure carrying these attributes. It has mainly four characteristics: Gender, product category, product descriptive text and size extension. Classification of products is done based on these attributes to create size grid names. Gender could be seen in the name as either men (M) or women (W). Product category can be either top (T) or bottom (B). T-shirts, shirts, hoodies are examples for top products. Jeans and shorts are examples of bottom products. Product descriptive text is a name decided by the merchandising team and it is special to the company. The name could reflect the target consumer group, the style and the fit of the product. Size extension is the fourth component of a size grid name and it was created mainly to denote how big the size grid matrix is. Low (L), medium (M) and high (H) are the examples. On the other hand, young (Y) for young consumers and Big{\&} Tall (B{\&} T) for overweight and tall consumers could be also some other size extension names. WB–Youth Super Skinny–M is an example of a size grid name where the products under this grid are for women bottoms with super skinny fit and the grid is medium in terms of number of sizes.\\

If a new size grid name is created, it will not have any history to make an analysis for size selection. For that reason, a new size grid name is designed based on product related insights and market data from global team and merchandising team. If a size grid name is an existing one, sizing planners conduct a historical analysis and make decisions for which sizes should be made available for a certain season(s). Sizing planners make the analysis on size grid name level. The main key performance indicator (KPI) of the decision making is adjusted demand. Other KPI’s are Sell-Out and Sell-Through. Adjusted demand basically shows customer demand in product quantities in a certain period of time. In order not to miss the quantities that customer demanded but was not met, it has a formulation of calculating the summation of both met demand which is the number of products sold to final customers and also unmet demand which is not met due to stock out cases or problems in supply chain and/or product planning. Sell-Out shows the sales that the company does to the final customer. It is basically what is sold in the stores. Sell-Out data is not available for all stores. The data is available only for the stores owned by the company and some stores owned by some planning groups such as Zalando. However, most of the planning groups, which are other companies Levi’s sells its products to, do not give or keep the Sell-Out data. Moreover, Sell-Through shows how well a product is performing. It is a ratio calculated by dividing the Sell-Out by initial inventory. Similarly, Sell-Through data is also not available for all planning groups because of both Sell-Out and Inventory data. Both are not available for all stores. \\

For the analysis, aggregation of the KPI’s for the last 3 or 4 seasons is taken for each size grid name and size. Normally, the aggregation used to be made for the last 2 seasons. However, due to the pandemic, the demand shows a fluctuating curve and in order not to risk stock out and excess inventory cases, a longer time horizon is taken under consideration. Size selection process is performed for both tops and bottoms for Europe region. In terms of size structure, tops mostly have more basic size structure than bottoms. Tops have one dimension in size while bottoms mostly have two dimensions where first dimension stands for waist and second dimension stands for length of the product. S, M, L, etc. are examples of sizes for tops and 32 30, 34 30, etc. are examples of sizes for bottoms. Decision making is based on size grid matrix. Non-selected sizes, the neighbors of which have high adjusted demand or other KPI’s has a potential for being a good performer and so being added into the size grid for the planned season. Removal of a size is similarly decided based on poor performers among sizes. Figure.27 in Appendix shows the size grid matrix and is an example of how a size grid name is evaluated for KPI’s. After the agreement between stake holders, the data with the final size selections is uploaded to the system. By that, all possible sizes for a product under a certain size grid name are decided for the production.\\

Further in the process, planners select the sizes from the grids for each planning group in the portfolio by using an internally developed tool. It is possible for a planner not to select a size which is already available in the grid. However, it is not possible to add more sizes to the available sizes in the grid. The ways of planner size selection is divided into two by sale channels: Wholesale and Retail. For the retail channel, planners make the same size selections for all planning groups. An upper limit can be set for the number of sizes to be selected by the planners. For instance, maximum 24 sizes limit was set for high (H) grids for Spring-Summer 2022. Retail planners takes the upper limit into consideration during the size selection. This decision is made by the sales directors because planners want to select as many sizes as possible in order to have more flexibility and to satisfy customer needs. However, wide range of available sizes can result in excess stocks at the end of a season, which is not desirable for the company. A new size is selected if the combination of Sell-Through and Sell-Out (or adjusted demand) of neighbor sizes is performing well. A neighbor size is defined as the closest size on same waist but different length or same length but different waist or both. If the Sell-Through of the neighbor sizes is high, a planner also checks Sell-Out quantities. Main idea is not to look at sizes with very high Sell-Through, however low Sell-Out. It is to add sizes that will help the company to sell more. A planner does not want to add sizes when they are not sure about its selling potential. On the other hand, wholesale planners who manage the sizes for the stores, which are not owned and managed by the company, make different size selections per planning group. They specially prioritize customers’ requests while making the selections. The customers can provide the information of which sizes they are going to order in the next season. In this case, size selection is done based on customer needs. For wholesale channel, there is no upper limit in number of sizes that can be selected like in retail. Customer request is more important for wholesale channel. The whole process flow of size grid building and management can be seen in Figure.1.\

\bxfigure{\textit{Size Grid Building and Management Process}  is managed by Sizing, Merchant and Planning Teams, where analytical tools are used for the support of data flow through the systems. SGMT is a frond end tool to import size selections into the database by the planners in the company}{\includegraphics[scale=0.75]{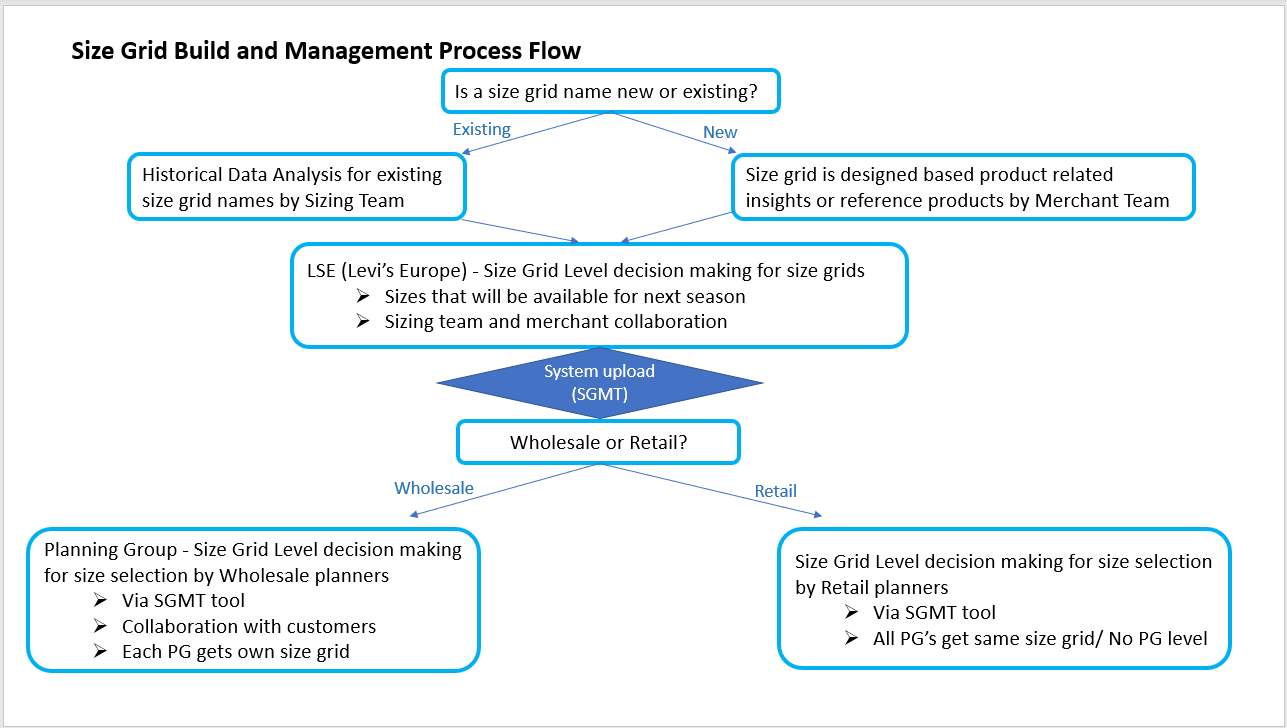}}

\addcontentsline{toc}{section}{2.2 Datasets}

\subsection*{2.2 Datasets}
In this section, some details of the datasets and the software tools which will be used in the study are given. The variables, how the data is kept for certain processes in the supply chain, the length of the data, how to merge them together to obtain a master dataset and how they will be processed in order to be used in the further steps of the study are explained. \\

Datasets are stored in the cloud storage in the company. Dataiku which is the platform for reaching the data and enabling to build artificial intelligence projects is used in the company. One project or multiple projects can be created in the software. For this study, a new project was created and, in the platform, mainly Python as a coding language and visual recipes designed within the tool will be used for data processing and model building. The platform also enables to create a project flow helping the user to follow the steps in a project. The flow of the study can be seen in Figure.28 in Appendix.\\

First dataset is for item plan forecast where it is fed by putting the forecast per product for certain months in a season by planners. It is basically denoting product level demand prediction in quantities and has ca. 11 million rows. Season logic is another dataset that shows the fiscal months for a product staying in the market. A season for different products can mean different time horizons even if they are planned for the same season. To illustrate, a Fall-Winter season is from June to November for core products while it is from May to December for seasonal ones. This table helps to differentiate the market life of different product groups and it has 790,979 rows. Size selection of products are kept in two different datasets. In Figure.1, it was mentioned that firstly sizing team builds the size grids and uploads to the system. Theses sizes are on size grid name level and valid for the whole European region. They are kept in Fact Sizes Available Conso dataset having 340,438 number of records. After the planners select the sizes among the decided ones in the previous step, it feeds Size Selections table where sizes on planning group level data can be found. This dataset has ca 6.3 million rows. In addition, there are two different datasets for keeping products and product related variables: Product Master and Product Conso. In the product master table, variables such as target consumer group, product category, product family, color and fabric exist. It has 618,045 rows. In the product conso, there are variables mostly related to the market such as first and last on floor, status, seasonality, destination. The table has 97,423 rows in total. Planning groups and their information about belonging to which channel and affiliate are in the planning group dataset with 30,258 rows. On a regular basis, planning groups are merged or remapped twice a year in the business. The mapping of the planning groups is kept not in the cloud but manually in a sheet which was uploaded to Dataiku. Its name is planning group mapping and has 336 rows in total. Each KPI used for decision making in size selection is kept in a different dataset. To begin with, adjusted demand historical dataset has ca. 8 million rows and shows the number of quantities sold and ordered for each planning group, season, month, size. Sell-Out is on product, size, planning group and season level data with ca. 2.7 million of records and showing the number of products sold in the stores. Sell-Through as a third KPI does not exist in a dataset. It is calculated based on Sell-Out and Stock datasets with the formula (2.1.1):\\

\hspace{1cm} Sell-Through = Sell-Out / (Sell-Out + Leftover Inventory)     \hfill                     2.1.1\\

Stock data is the biggest data set with ca. 100 million rows among the other datasets that will be used in this study. Similar to Sell-Out data, it is on size, product, planning group and season level and showing the left stock end of the period in quantities. Although it is the biggest one, it has many null values in season column. A summary of the tables and explanations can be seen in Table.6 in Appendix.\

\addcontentsline{toc}{section}{2.3 Data Cleaning and Processing}

\subsection*{2.3 Data Cleaning and Processing}
In this section, cleaning of the datasets and processing them in order to obtain a master dataset to be used in model building are explained. \\

Previously, it was explained that the size grid building and management is done on size grid name level and not on product level. Since the datasets are mostly on product level, a proper aggregation is needed so that the datasets will be merged on size grid name level. Size grid name of a product can change through the seasons. While some products can be removed from a certain group, some can be added or a size grid name can be canceled fully. Following these changes, products need to be mapped to size grid names in a way that evaluation of size selections should not be impacted when model is trained with the historical data. For this purpose, size selections and fact sizes available conso datasets were merged with the variables: product code, season and size grid name. By that, all size grid names created in the company in all seasons in the history was obtained. Additionally, product master data, which has product category information, was merged to this dataset. Size grid name and product mapping was done only for product categories tops and bottoms. Mapping was performed by sorting product codes and size grid names by the season and taking the size grid name of the last season for each product. The logic comes from the tool called as AIO, where sizing planners take data from to make the analysis of size selections for each grid name. After the mapping, the obtained dataset has 21,354 rows and the information of unique product codes and mapped size grid name, and this dataset will be used for other datasets to map the product codes to size grid names.\\

The dataset for planning groups has 1 missing row and it was removed. For the rest, the variables do not have any missing data. However, some planning groups are duplicated because of the organizational structure and the brand names. There are two channel planning directors and two brand names (Dockers and Levis) under each planning group. For this study, the necessary variables from this dataset are planning group name, channel and affiliate name. After removing the duplicate records, the number of rows went down to 49. \\

Stock data is the biggest data with ca. 100 million rows. However, the table has a lot of missing values for some variables. 98\% of the season, 5\% of the size and 3.8\% of the planning group columns are missing. As the analysis is done on planning group, size and also season level, the missing values cannot be ignored or imputed and need to be removed from the dataset. After removing the null values for these columns, 1,145,908 rows were remained. Size as a variable is a string column and it was processed to get rid of special characters such as colons, dashes and blank characters. Size grid name and product mapping table was merged with the stock data to add size grid name column to the data. Afterwards, the stock quantities are aggregated on size grid name level. Sizing planners are making the analysis based on a couple of seasons while checking the KPI’s. Normally, aggregating KPI’s for 2 seasons would be enough to decide which size to add/ drop in the grid. However, due to the pandemic conditions, the demand is fluctuating, which has made the selection more difficult. For that reason, KPI values were aggregated by taking  the recent 4 seasons summation for each record. For instance,  adjusted demand of season 193 (Fall-Winter 2019) reflects the adjusted demand of the previous 4 seasons: 191, 183, 181 and 173. There are two seasons in a year. Season code is structured in a way that first two character represents the last two digits of a year and the last character represents the half. If the last character is 1, it means Spring-Summer. If it is 3, the season is Fall-Winter. After aggregating on size grid name level, the dataset with 149,310 rows was obtained. Mainly, size grid name, size, planning group, season and the stock quantities are the columns in the final dataset.\\

Among the variables of the Sell-Out data, 6\% of the size, 2.7\% of the planning groups and 51\% of the season columns were missing. Similarly, these columns are required in further steps in order to join the tables. Unfortunately, the columns with missing values need to be removed. After filtering out the rows with missing values for one of the columns, the size column was processed in order to get rid of the special characters similar to the stock data. Sell-Out data was merged with the size grid name mapped table on product codes. The sold quantities are aggregated on size grid name level, and the data set with 165,246 rows was obtained finally.\\

By using Sell-Out and Stock datasets, Sell-Through, which is the third performance indicator for size grid building and management, was calculated. Before joining the two tables, it was seen that Sell-Out data had negative values for the sold quantities where it can only be zero or a positive number. Sell-out represents net sales and there are some cases where the number of returned products is higher than gross sales. These cases are causing negative values and they are mainly coming from e-com customers. Stock shows left stock after sales. Negative values for stock is caused by that it is possible to sell more than the stock within the system although it is not possible in real. Negative values are noisy data and were removed from the data. Then, Sell-Out and Stock datasets were merged on size grid name, season, size and planning group level. Before the calculation of Sell-Through by using stock and sell-out, the rolling 4 seasons for stock and sell-out quantities are calculated. Afterwards, for the calculation of Sell-Through, a new column was created and the formula in 2.1.1 was used. The calculation was made by using a condition where the summation of stock and Sell-Out cannot be zero as it is the denominator in the division and its zero values make the Sell-Through ratio infinitive. Thus, Sell-Through values are left as null values when both stock and Sell-Out are zero in the row. Finally, a dataset with 165,246 rows was obtained with the two KPI’s, Sell-Out and Sell-Through on planning group, size, season and size grid name level.\\

Adjusted demand dataset is already a processed one. The raw data is sent from global and it is run and refreshed weekly. The job was created and is managed by Artificial Intelligence team. Despite its being processed, missing values and further processing was performed for this study. The expectation for this table is having fewer missing and noisy data because adjusted demand as a KPI is the main one and available for both sales channels unlike Sell-Out and Sell-Through. The missing values was checked and the variables which are used in the study don’t have any missing data. On the other hand, the data is stored on month level. Hence, the adjusted demand was aggregated on season level by planning group, product, season and size. After the aggregation, it was merged with size grid name mapping dataset on product code and aggregated adjusted demand once more on size grid name level. As a last step, the recent four seasons’ demand data was taken on a rolling basis similar to Sell-Out and Stock datasets. The final adjusted demand data with 345,412 records was obtained.\\

Item plan forecasts are on product, planning group and month level. Before each season begins, planners make their forecasts for all months in the season and upload to the system. Although full season forecasts are prepared before the season, the numbers can be updated monthly. Putting a forecast for a product on a certain planning group and month decides the assortment of the products in the market. For each product, a season could mean a different time horizon depending on the product type. In order to know which product in which month belongs to which season a dataset called season logic is stored in the cloud system. The season logic dataset has material, month and season information. Item plan forecast dataset was joined with the season logic on product code and month so that product and season information was obtained in a unique dataset. Afterwards, size grid mapping table was joined to the table. The obtained dataset is on product and month level where size grid name and season level is needed. Before removing the product and month variables, in order to filter out some variables product master dataset was joined to the table. From product categories, tops and bottoms are filtered and accessories and footwear products are removed. Additionally, outlet products are not in the scope of this study. They were removed from the dataset. There were still some product codes which are not actual product codes and called as dummy products. They were also filtered out from the dataset. As a last step, planning group, season and size grid name as variables are selected, duplicated records were dropped and a dataset with 40,034 rows was obtained.\\

Size selections are stored in two different datasets. In order to obtain a dataset having historical size selections on size grid name, planning group and season level, firstly size selections that were made via SGMT tool by the planners were read from the table. The sizes which were selected before and dropped later have a status ‘D’. Dropping a size could be caused by the wrong selection, changes in size selection decision strategies or any changes made in assortment during planning. Dropped sizes were filtered out from the data. The size selection process in the company is if there is a size selection for a certain product in SGMT tool, the selections are made among the sizes that were created during size grid building. Normally, if a planner wants to select fewer sizes than the grid or wants to make special size selection for certain planning groups, the tool is helpful. If all sizes that exist in the size grid need to be selected or there is no need to make special size selection for planning groups, the planner could skip making the size selection via the tool. In that case, all sizes in the grid will automatically be selected for the products under a size grid name according to the assortment. These sizes are stored in a different dataset on size grid name level, with no planning group information, meaning that the sizes are on affiliate/ channel level. Thus, joining the tables should be firstly merging the size selections made in the tool. Afterwards, planning group, size grid name and season combinations which are remained without sizes needed to be filled with the sizes at affiliate/ channel. For that reason, the size selections that were made in the tool were initially joined to the assortment data. Season, planning group and size grid name combinations that were remained with no sizes after the merge are taken from the data and joined with fact sizes table. Then, both tables were concatenated and historical data which the size selection for all seasons, planning groups and size grid names was obtained. After the concatenation, the size column was cleaned by removing the special characters such as comma, column and blanks in the string and the data was written as size selected data into a dataset.\\

For size grid building and management, the decision to make is whether to select a size or not for a size grid name. For all the seasons in the history, sizing planners and merchandising team made the decisions for each size. However, the available data is only for selected sizes, not all candidate sizes. This situation brings the need of having all candidate sizes to evaluate for selection so that the model can be trained accordingly. For that reason, all candidate sizes for each size grid name should have been created in a dataset. However, one challenging task is that sizes in the grids do not have a unique format. Considering the bottoms having two dimensional sizes as waist and length, while the waist dimension could be consecutive numbers till a certain waist size and then could be only even numbers. For instance, one of the core size grid names and the waist dimension for the grid has consecutive numbers between 28 and 34 for Summer-Spring season. After 34, the waist size takes only even numbers till 44, meaning that there are no 33, 35 odd sizes that were planned to be produced (Figure.2). In fact, the format can change from one size grid name to another. There is not a commonly accepted size grid structure for any size grid name in the business. Hence, the automated data creation for which candidate sizes to be evaluated in a size grid name was not possible. Firstly, a dataset having all size grid names in the history and all possible candidate sizes  was manually created. For the dataset, two columns were created for sizes as first and second dimension. While creating the data, since there is no official upper or lower limits for the sizes, normally it was possible to create as many sizes as possible for the evaluation. However, considering the number of the selected sizes, adding too many candidate sizes which were not selected in the history could create very imbalanced dataset in terms of the response variable. For that reason, this was taken into account while creation of the candidate sizes. The created dataset with all possible candidate sizes had 658 rows, meaning 658 unique size grid names, and there are size grid name, size first dimension combinations and second dimension combinations as columns. \\

\bxfigure{\textit{Size Grid Matrix}  is a two dimensional grid example where adjusted demand is filled in the cells. The grid is for a certain size grid name in one of the Spring-Summer Seasons. The values are in product volumes}{\includegraphics[scale=0.75]{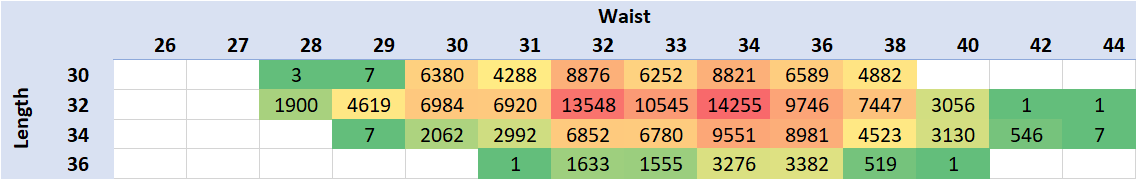}}

As a next step, selected sizes data was merged with all possible sizes data in order to take planning group and season information. Then, all possible size data was transformed by using the Cartesian products of first dimension and second dimension size columns in a way that each size combination was written in a new row with other columns. A new column with an integer value 1 was created in selected size data where the column shows the size was selected and after it was joined with all possible sizes dataset. After having this column in all possible sizes dataset, null values were filled with zero meaning that size with zero values were not selected for a certain season, planning group, size grid name combination. By that, the response variable was created. It is a binary variable and showing whether a size was selected or not. The data with selected and not selected sizes has 2,344,688 rows.\\

For further data processing, the dataset was partitioned by planning groups, where there were 39 partitions in total. The aim of partitioning the data was that the dataset became large and in order to process and create new attributes, parallel machines were needed for data processing. By that, processing time decreased and it helped to come up with a highly efficient process in terms of run time. After having obtained the partitioned data, records belonging to season Summer-Spring 2021 and later seasons were filtered out since datasets for these seasons were partially complete or not complete at all. When the data with inclusion of selected and not selected sizes history was obtained, all processed datasets were merged to the assortment dataset. Thus, if data did not exist in one of these datasets, they became null values based on the variables used for joining the tables. For adjusted demand, null values were filled with zero since adjusted demand data was available for all sales channels. It means when there is no data for adjusted demand, the product and size for certain planning group in a season was not sold any. Unfortunately, this situation was not valid for other KPI’s. As previously mentioned, stock, Sell-Out and so Sell-Through datasets were not stored for some planning groups as they were not available or provided by the customers. Therefore, null values for these KPI’s were filled in a different way than adjusted demand. For the sizes which were not selected in any seasons, it was obvious that the values can only be zero for stock, Sell-Out and Sell-Through variables. The null values complying with this condition were filled with zero. On the other hand, it was unknown whether the values are zero or missing for the remaining null values of stock, Sell-Out and so Sell-Through. The reason is that a data is not stored if there is no sales and not available data could also be caused by that the data cannot be provided by the partners. Since the case is different than surely knowing the values of variables are zero, 50 new dummy binary variables were created for stock, Sell-Out and Sell-Through. 1's for the dummy variable represent the data is missing and 0's are for not missing data.\\

After having handled the missing values for adjusted demand, Sell-Out and Sell-Through, creating new attributes for taking the neighbor values of adjusted demand, Sell-Out and Sell-Through for a size in the size grid was the next step. The decision made by sizing planners and merchandising team for size selection is done by evaluating addition of a candidate size by looking neighbor size performances. For the size selection decision making process, please see Figure.1. If a candidate has neighbor sizes in the grid performing well in terms of the KPI’s, it is possible that the candidate size can perform well, too. Size planners decides addition of the candidate into the grid. For example, a candidate size in a size grid can be seen in Figure.3. In the example, the candidate size has waist 29 and length 34. Performances of all neighbors of this size, which are 28 32, 29 32, 30 32, 28 34, 30 34, 28 36, 29 36 and 30 36, are normally checked by the planners while decision making. In the example, it was seen that neighbors 29 32, 30 32 and 30 34 are good performers which makes the candidate size 29 34 a possible good seller. For this reason, this size was added into the grid during the season planning. Consequently, considering each size in the grid has 8 neighbor sizes, 8 new attributes are added into the dataset showing the neighbor performances in the first circle. By that, the model will consider the neighbors of a size while deciding whether to select a size or not. However, although it is not common to add a size where none of the neighbor sizes have any historical data that shows the size performances, it is still possible to add a size lying in the outer circle of the selected candidate if the size is promising for its performance in the future. Waist 27 and length 34 is such an example in Figure.3. The neighbors around the sizes have no data. In this case, model cannot learn to select this size by taking performances into consideration. Therefore, in order to integrate the neighbors lying further from a size in the grid, the neighbors in the second circle in the grid were added as new attributes into the data set. The neighbors in the second circle of a size 29 34 are 27 30, 28 30, 29 30, 30 30, 31 30, 27 32 31 32, 27 34, 31 34, 27 36, 31 36, 27 38, 28 38, 29 38, 30 38 and 31 38 in the example seen in Figure.3. With these additions, there are 24 new attributes for each KPI. \\

With the addition of neighbor size performances into the dataset, the model will be able to take the decision making of size selection by looking the neighbor sizes into consideration. One aspect of size selection is if the size has no historical performance, neighbor sizes become important and neighbor sizes were initially added into the dataset without weighting the values. However, the further a neighbor size is from a candidate size in the grid, the less impact it should have on the decision making of size selection. For that reason, a weighting method for performances was considered. Manhattan distance  was considered to calculate the distance between the candidate and its neighbors. A distance matrix was created and the calculated Manhattan distances were used to give weights for KPI performances. The idea was dividing the KPI's of neighbor sizes by Manhattan distance matrix to obtain weighted performances based on the neighbor locations. Like a candidate in Figure.3, 29 34 is 1 unit distance from its neighbor size 29 32. Since the division by 1 unit will give the same value, a candidate size was assumed to be in location where it is 1 unit distance away from a theoretical point. Manhattan distances for all other neighbor sizes were calculated based on this theoretical point. Thus, if a neighbor size is 1 unit far from a candidate size, its distance was calculated as the distance from the candidate size plus 1 unit distance, where it makes 2 unit distance for this example.  \\

\bxfigure{\textit{A candidate size} can be defined as a size to be added or removed for the following season. The evaluation is done by analyzing the neighbor performances when the size has no history. x-axis shows waists and y-axis shows length as size of a product}{\includegraphics[scale=0.75]{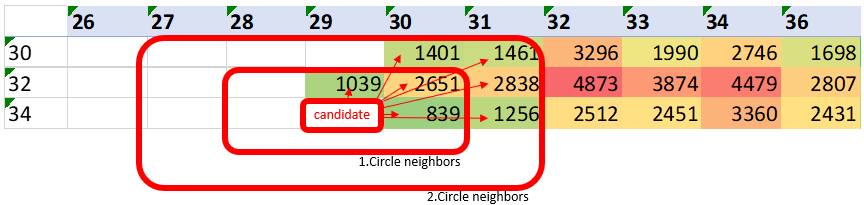}}

The company has two brands under its name, Levi’s and Dockers. Levi’s is the driving brand in the business and have a bigger share than Dockers in terms of selling quantities and revenue generation. That makes most of the data for Levi’s brand. The decision makers in the company requested to focus on only Levi’s brand. At this point, the data will be divided into two by the brand name and further analysis and model building will be only for Levi’s brand. The processed data for Dockers will be stored in a different dataset. In case, the decision changes in the future, all codes that were written for Levi’s can be used for Dockers as well. After the division, the data for Levi’s brand was remained with 1,659,018 rows.\\

\addcontentsline{toc}{section}{2.4 Univariate and Multivariate Analysis}
\subsection*{2.4 Univariate and Multivariate Analysis}

 In this section, the variables in the master dataset were analyzed by univariate and multivariate analysis. There are both categorical and continuous variables in the data. Both were analyzed as candidates in model attributes. The master dataset which was obtained in the last step has 1,659,018 rows in total.  \\
 
 Among the continuous variables, other than Sell-Through variables, other variables’ units are product quantities. Sell-Through is a ratio that can take values between 0 and 1. Mean $\pm$ standard deviation for adjusted demand is 178$\pm$2,460, for Sell-Out is 83$\pm$1,743, for Sell-Through is 0.04$\pm$0.15 and for stock is 41$\pm$534 (See Figure.4). Min and max values that adjusted demand can take are 0 and 313,406, that Sell-Out can take are 0 and 387,727, that stock can take are 0 and 105,753 and that Sell-Through can take are 0 and 1. 0.25, 0.50 and 0.75 quantiles are zero for these four variables (See Figure.4), indicating there are too many zero values in the columns. This can be driven by two reasons: selected sizes were not sold or the data for not selected sizes. Moreover, the standard deviations of the variables are much higher than their mean values. The situation is natural in the fashion and apparel industry. Some products are volume drivers in sales and there is very high customer demand for them. On the other hand, depending on the product, size, planning group, an SKU could have very low sales in quantities. The other continuous variables in Figure.4 are the neighbors of a size in the grid for the same KPI’s. Since they were created by using existing variables, they have similar behavior in terms of very high standard deviation comparing to their mean values and too many zero values.

 \bxfigure{\textit{Mean, Standard Deviation (SD), minimum value, 25, 50 and 75 percentiles, maximum value of variables} are presented. Standard deviations are relatively big compared to the mean values. Zero percentiles are caused by many zero values in the variables}{\includegraphics[scale=0.65]{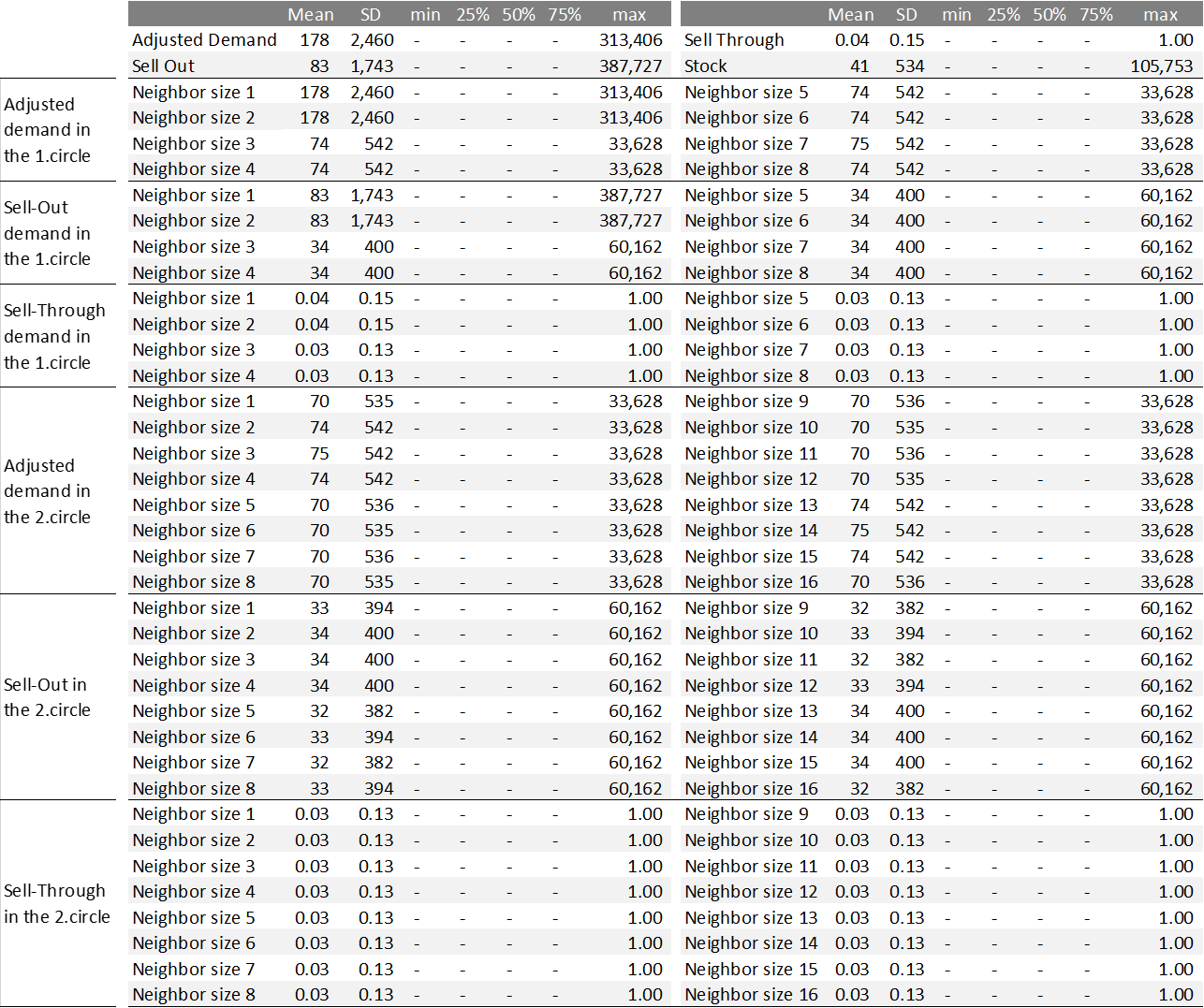}}
 
 Among the categorical variables, there are 9 different columns in the dataset: size grid name representing a group of products having similar characteristics, size first dimension which can be waist for bottoms or the size of a product for tops, size second dimension which can be length for bottoms or null values for tops, planning group that shows the customer group(s) products sold to, seasonality LSE which shows whether a product is new, seasonal or core, channel stating sales channels as wholesale or retail, affiliate name stating which affiliate in Europe region the company is operating, gender showing the consumer group as males or females and product category as bottoms and tops (Figure.5). Unique row in Figure.5 shows how many levels each categorical variable has. There are 464 levels for size grid name, 79 levels for size first dimension, 26 levels for size second dimension, 38 levels for planning group and 10 levels for affiliates. For other categorical variables, there are 3 levels for seasonality LSE, 2 levels for channel, 2 levels for gender and 2 level for category. It is important to note that the number of levels for some variables are high which is important for modeling and will be discussed in model approach section. Among size grid names, MB-511-H is the most frequently seen grid name, which is for men bottoms, 511 fit, high size grid products. 28 from size first dimension and 30 from size second dimension are the sizes that have the highest frequency among others. South Field Accounts which is a group of customers located in Southern Europe is also the mostly seen in the data among planning groups. Core for seasonality LSE, wholesale for channel, Central for affiliate, male for gender and bottoms for category are the other mostly frequent variables in the master dataset (See Figure.5). Having the high frequency does not necessarily mean that they are highly demanded by the consumers. It means these groups have more combinations in the data than other groups. However, it could also be an indication of being volume driver in the variable. \\

 \bxfigure{\textit{Categorical Descriptive Statistics:} High number of levels for some categorical variables are remarkable}{\includegraphics[scale=0.65]{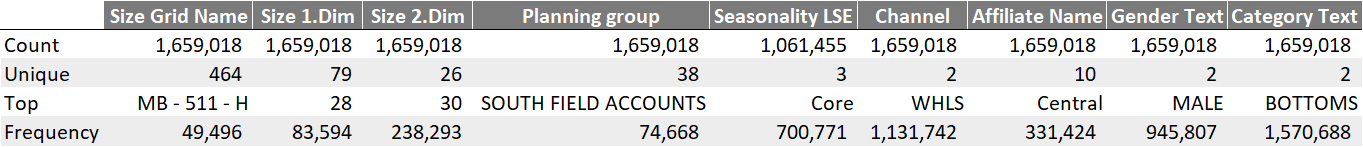}}

The distributions of continuous variables were checked by plotting histograms with smooth curves. It was seen that all variables have zero-inflated curves. There are too many zero values for the variables. The plot can be seen for adjusted demand in Figure.6. The remaining variables have similar curves. In order to see the distribution of the continuous variables for nonzero values, the histograms were plotted additionally by excluding zeros. Adjusted demand, Sell-Out and the same variables for the neighbor sizes have very right skewed curves. Adjusted demand for neighbor size 3 in the first circle can be seen in Figure.7. The curves show that the observations are highly populated in small values relative to very high values on the right tails. The mean value of this variable was 81 while the maximum value is located at 33,628 which is very far from the mean (Figure.4).

 \bxfigure{\textit{Adjusted Demand} distribution is populated at and around zero despite heavy tail on the right}{\includegraphics[scale=0.85]{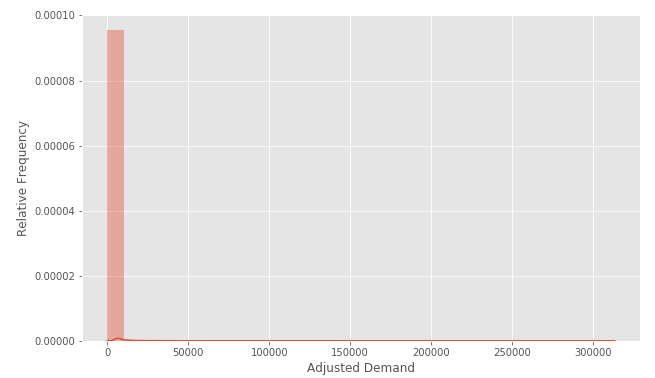}}

 \bxfigure{\textit{Adjusted Demand for neighbor size.3} histogram plot with smoother (exluding zeros) is clear to see the distribution as very right skewed}{\includegraphics[scale=0.85]{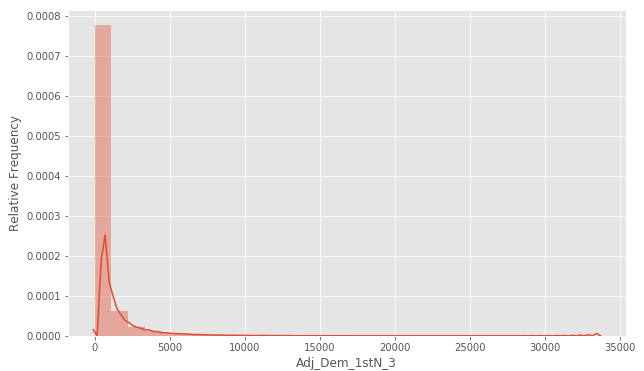}}

Unlike adjusted demand and Sell-Out curves, Sell-Through has a bell-shaped curve and more centrally distributed for its nonzero values, slightly skewed to the left (Figure.8). It was seen that the mean value was near 0.6 for the nonzero values, meaning that 60\% of the inventory was sold on the average for nonzero observations. On the other hand, Sell-Through ratio curve for neighbor sizes is a bit different. The observations are populated near the values which are a bit higher than zero. The other observations have also bell-shaped curve between 0.1 and 1 on x-axis (Figure.9).The distributions of other KPI's and their neighbor sizes in 1.circle for nonzero values can be found from Figure.33 to Figure.55 in Appendix. For the purpose of more explanatory data analysis, especially for outliers, box plots were plotted for the variables without zero values. Considering mean and max values the variables can take, there are too many outliers for adjusted demand, Sell-Out and the same performance variables observed for neighbor sizes as expected (Figure.10). Outliers for the variables have high values and the outliers are representing volume driving highly demanded products. Therefore, they should not be removed from the data. In fact, they are the most significant products in the assortment for the company. The box plots of other KPI's and their neighbor sizes in 1.circle for nonzero values can be found from Figure.56 to Figure.82 in Appendix. The other variable which is the result of decision making for size selection is size selection status. It is considered as the response variable for the model and called as size selection status. It takes 1 if a size is selected. Otherwise, it is zero. 1,277,620 rows are zero and 381,398 are one for the variable (Figure.11). It means 381,398 sizes were selected in the history while 1,277,620 were not. Like other continuous variables, the size selection status also has too many zero values. In fact, number of zeros is three times more than one values where it was seen imbalance between the two. Imbalance in the response variable could cause poorly trained model which will be discussed in model approach chapter. Alternative ways of dealing with imbalance class variable will be applied.\\

 \bxfigure{\textit{Sell-through} plot shows a distribution closer to the normal distribution with skew to the left}{\includegraphics[scale=0.85]{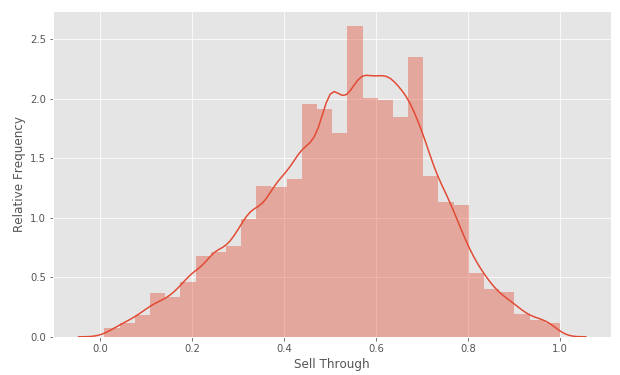}}

 \bxfigure{\textit{Sell-through for neighbor size 1 in the first circle} (excluding zero values) gives a distribution with heavy tails on the right}{\includegraphics[scale=0.85]{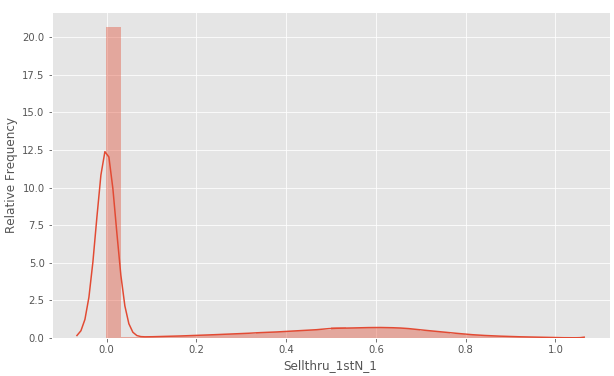}}

 \bxfigure{Boxplot of \textit{Adjusted demand for neighbor size 3 in the first circle} (excluding zero values) indicates the existence of many outliers for the variable. High variances from the median for many outliers are remarkable }{\includegraphics[scale=0.85]{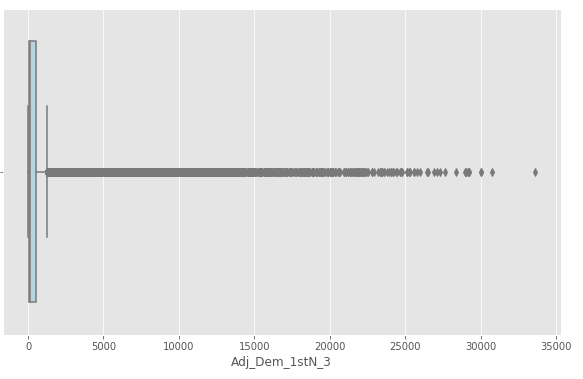}}

\bxfigure{\textit{0: Size is not selected /
1: Size is selected} \\ Non selected sizes dominate selected sizes in terms of frequency}{\includegraphics[scale=1.3]{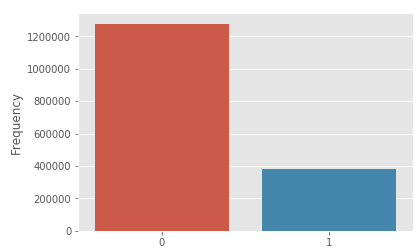}}

Checking the number of size grid names in the historical data, it was seen that there were only female size grid names kept in the data in Fall-Winter 2016, the first season in the history. It does not necessarily mean that there were no size grid building and management for males in Fall-Winter 2016. It could be whether the data hadn't been kept yet or the size selection was being processed differently than today. Through the seasons, number of size grid names for males increased till Fall-Winter 2018 (183) and it almost doubled the number of size grid names for females in 183 (Figure.12). After the season 183, it started to decrease and the number of size grid names for both genders became almost equal in 203. Merchandising team prefers to have as many size grid names as possible to have more flexibility in terms of the products in the market. On the other hand, sizing analysts and planners want to make it more structured and as few as possible for operational efficiency and analysis purposes. This was seen in the number increasing for male grid names in time and with the actions they were merged, and fewer grid names were left available for the decision making in the recent seasons. As the company grows in time, more products and different product types have been made available in the assortments. This naturally affected the change in the numbers of size grid names in time. For females, it was around 40 in 2016, while the number was above 80 in 2020.

\bxfigure{\textit{Number of size grid names} follow different trend through the seasons by gender while the balance between the genders has been obtained recently}{\includegraphics[scale=1]{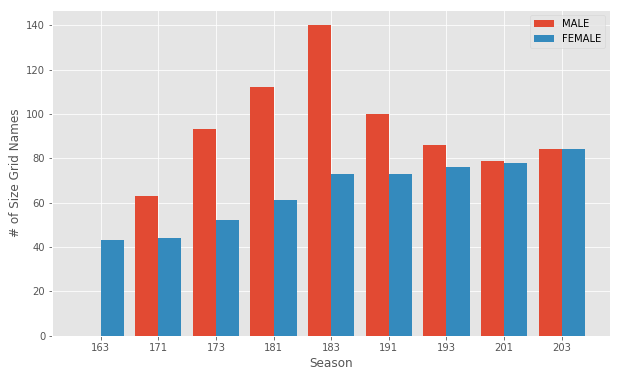}}

Adjusted demand is the main KPI used for the size selection analysis and the analysis is done by looking the last couple of seasons’ aggregations of the demand. Figure.13 shows the adjusted demand in the history by gender where it was obtained by aggregating the recent four seasons’ adjusted demand. Adjusted demand was seen higher for the products designed for males than the products for females in all seasons in the history. From the first season to the last season in the data, the adjusted demand shows an upward trend for both genders. After the season 191, the demand had a sharp increase till the beginning of 201. The increase in adjusted demand between the seasons went down from 201 to 203 comparing to the previous two seasons. This was caused by the pandemic of Covid-19. With the pandemic, the demand of the products showed a downward trend after having closed the stores in many countries in Europe.\\

\bxfigure{\textit{Rolling adjusted demand} has lost its acceleration in Fall-Winter 2018 and Spring-Summer 2020 (caused by the pandemic)}{\includegraphics[scale=1]{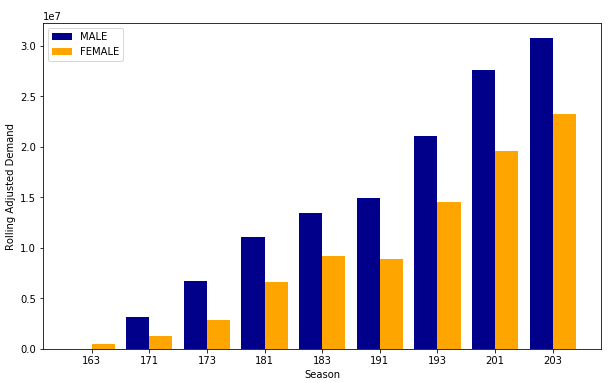}}

The KPI’s were also analyzed in order to see highly demanded sizes for each product category. The size variable was split into two dimensions. The first dimension is showing waist for bottoms and the size of a product for tops. On the other hand, the second dimension shows length for bottoms. The second dimension for tops is null. Among the bottoms, there are also products with one size dimension. They can be produced with a fixed length for all waists or the size of a bottom product can also be designed like tops as one dimension in size for some special products. The size charts were designed for each gender and product category in the company as the anatomic measurements are different between males and females (See Figure.29, Figure.30, Figure.31 and Figure.32 in Appendix). 34, 32 and 33 are respectively the waists that male customers purchase the most from bottoms (Figure.14). Waists 34 and 32 are very close to each other in terms of the demand while the adjusted demand for waist 33 is relatively lower comparing to them. On the other hand, 27, 28 and 26 are the women waists with the highest adjusted demand respectively for bottoms (Figure.15). The figures are in line with the results of the study conducted in 2007 by Flegal KM [7]. Men have bigger anatomic measurements for waist than women on the average although the distribution of waist for both genders coincides for people at different ages. \\

\bxfigure{\textit{MALE BOTTOMS - Size 1.dimension: Waist for a bottom product vs Adjusted Demand in units} 34, 32 sizes are highly demanded sizes for male bottoms}{\includegraphics[scale=0.85]{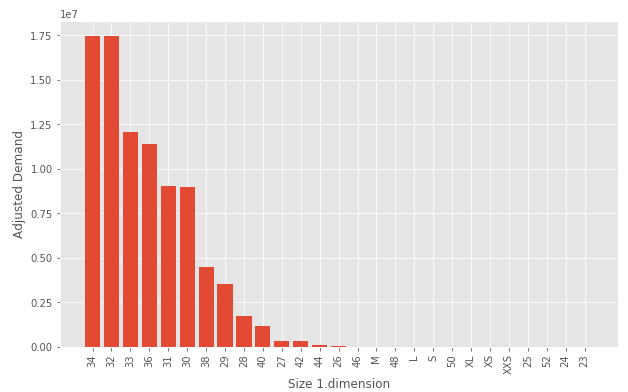}}

\bxfigure{\textit{FEMALE BOTTOMS - Size 1.dimension: Waist for a bottom product vs Adjusted Demand in units.} 27, 28 sizes are highly demanded sizes for female bottoms}{\includegraphics[scale=0.85]{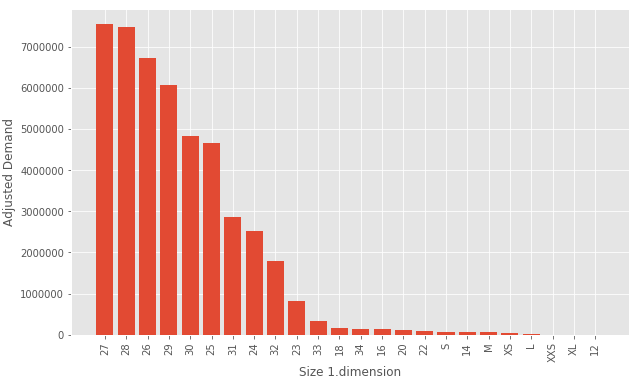}}

Similarly, the ranking for the length in size is respectively 32, 34 and 30 for male consumers (Figure.16). Length 32 has relatively high adjusted demand among the other lengths. In fact, the demand for length 32 is more than twice the length of the second highest demanded length which is 34. For also females, length 32 has the highest adjusted demand in bottom products (Figure.17). However, corresponding lengths in cm of length 32 are different for male and female. Length 30 comes next for females with the second highest adjusted demand. The other sizes for length in female products are relatively small values comparing to lengths 32 and 30. Comparing men and women in terms of length sizes, adjusted demand for men is distributed to 4 sizes while it is distributed to 10 sizes for women. This shows female consumers being more sensitive than male consumers in terms of sizes while purchasing a bottom product. \\

\bxfigure{\textit{MALE BOTTOMS - Size 2.dimension: Length for a bottom product vs Adjusted Demand in units.} 4 sizes are driving the sales volumes for male bottoms}{\includegraphics[scale=0.85]{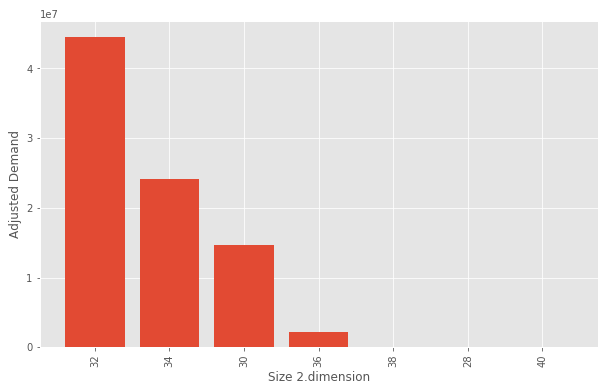}}

\bxfigure{\textit{FEMALE BOTTOMS - Size 2.dimension: Length for a bottom product vs Adjusted Demand in units.} More size are available for females than males in bottom products}{\includegraphics[scale=0.85]{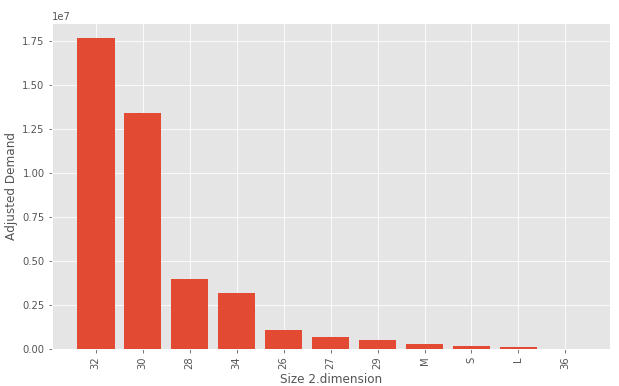}}

Tops have smaller number of size combinations than bottoms in general. While a bottom product can take up to 40 different sizes, a top product takes 5 to 6 sizes in general. This makes the size selection decision making easier for tops. The top 3 sizes for male tops are M (Medium), L(Large) and S(Small) in terms of adjusted demand in order (Figure.18). On the other hand, the top 3 for female tops are respectively S, M and XS (Figure.19).\\

\bxfigure{\textit{MALE TOPS - Size 1.dimension: Size for a top product vs Adjusted Demand in units.} M and L are the most demanded sizes for male tops}{\includegraphics[scale=0.85]{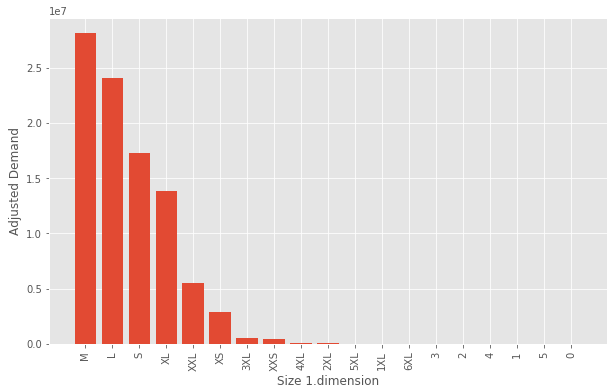}}

\bxfigure{\textit{FEMALE TOPS - Size 1.dimension: Size for a top product vs Adjusted Demand in units.} S and M are the most demanded sizes for female tops}{\includegraphics[scale=0.85]{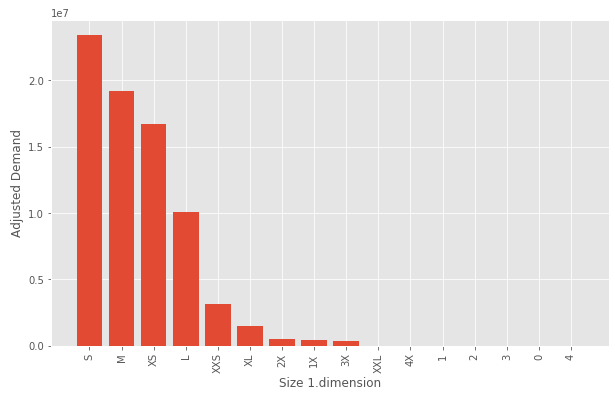}}

The company groups the customers, that it sells the products to, under planning groups depending on the geographical locations of the customers, business needs, supply chain structure. A planning group does not mean the stores where the company ship the products to. The orders of the planning groups are met from four distribution centers located in different countries in Europe. Although there is an allocation methodology in the business, a distribution center still can send the products in the stock to the customers which normally it does not serve when it is needed. South Retail has the biggest share with 12\% in adjusted demand for the company as a planning group (Figure.20). The planning group is one of the volume drivers and so very important in the company’s portfolio. Top 6 planning groups cover more than half of the adjusted demand of the company for Europe region (52\%). The other five planning groups are South Field Accounts, Central Retail, North EU Dept Stores, North EU Field Accounts and Zalando. Based on the adjusted demand aggregated for the last four seasons, southern and northern countries are driving the sales.\\

\bxfigure{\textit{Top 6 planning groups}(South Retail, South Field Accounts, Central Retail, North EU Dept Stores, North EU Field Accounts and Zalando) are dominating others in terms of sales for both genders and product categories}{\includegraphics[scale=1]{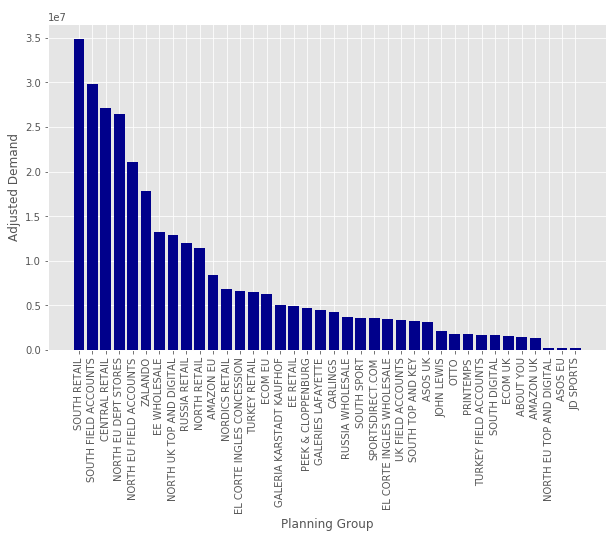}}

Adjusted demand was analyzed for the other categorical variables. For two sales channels as wholesale and retail, wholesale had a share 61\% while retail channel had the remaining 39\%. In terms of gender, male consumers took a share of 59.8\% from the pie which was 19.6\% higher than females. For the product categories, the adjusted demand is almost balanced between bottoms and tops where tops were with the share of 55\% and bottoms were with 45\%. There are 10 different affiliates mainly representing a country or a region. Central as one of the affiliates took 29\% of the adjusted demand and comes first among other affiliates (Figure.21). France followed central with 25\% of the demand. North with 12\% and LSE HQ with 11.7\% are the third and fourth affiliates having the biggest share in the chart. LSE HQ is the short version of Levi’s Europe Head Quarter, and it mainly stands for E-Com where the company sells its products online.\\

\bxfigure{\textit{Adjusted Demand} is shared among 10 affiliates, where Central, France and LSE HQ are with the highest shares respectively. LSE HQ stands for levi.com as the online sales platform of the company }{\includegraphics[scale=1.5]{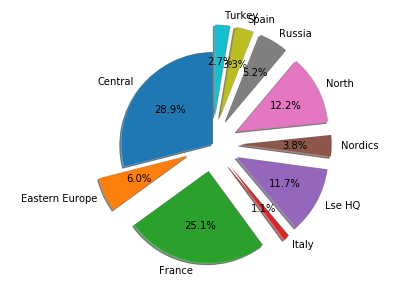}}

In order to check the statistical relationship between the continuous variables, the correlation matrix was created (Figure.83, Figure.84, Figure.85, Figure.86 and Figure.87 in Appendix). Adjusted demand is highly correlated with sell out, adjusted demand of neighbor size 1 and adjusted demand neighbor size 2. The correlation coefficients are respectively 0.7, 0.8 and 0.8. It’s correlation with both sell out neighbor size 1 and sell out neighbor size 2 are 0.6. The correlation between the adjusted demand and the other variables are relatively low. Sell out is highly correlated with sell out neighbor size 1 and sell out neighbor size 2. The correlation coefficient for both is 0.8. As the third KPI, sell-through is highly correlated with sell-through neighbor size 1 and sell-through neighbor size 7. These high correlations can be expected by the nature of decision making for size selection. As previously mentioned, a size with a high performance have tendency to have neighbor sizes with high performances. To sum up, it was seen that there are highly correlated variables in the data, and it needs to be taken into consideration in model approach.\\

\addcontentsline{toc}{section}{2.6 Methods}
\subsection*{2.6 Methods}

In this section, the modeling approach to the problem and applicable models were mainly discussed with their assumptions and limitations. Additionally, data split into train, validation and test sets, the response variable, the attributes, scaling and encoding attributes were explained. As the decision making for size grid building and management is based on selection or deselection of a size in the size grid, the response variable was selected as whether a size is selected or not. If the decision is to select a size, the variable will take 1. Otherwise, the decision will be not to select it and the response variable will be 0 (See 2.6.1). There are 75 continuous and 9 categorical variables which will be used as the attributes in model building in addition to 50 dummy variables that were created for missing data handling. Differences in magnitudes for the continuous attributes can cause some features with large values dominating over the features with small values. For instance, adjusted demand can take very large numbers while sell through can take values between 0 and 1 (See Figure.4). This can result in undesirable influence for the model results. Standardizing the features is a common necessity for machine learning methods. It helps the attributes to be treated equally in the model training [8]. Standardizing typically works by removing the mean of the variable and scaling it to unit variance. However, the variables in the dataset have many outliers where the sample mean/ variance can be affected in a negative way. One of the scaling methods called as Robust Scaler can deal with the outliers and give robust statistics to outliers [9]. Instead of using regular formulation for scaling, it uses the median and interquartile range while scaling a variable. This method gives better results for variables with many outliers. However, the variables in the dataset also have many zero values. Having too many zero values makes the median values, 25\% and 75\% quantiles zero. This makes the robust scaler unfunctional for our dataset. Therefore, standard scaler, which is standardizing them by removing the mean and scaling out to unit variance, was used for all continuous variables [10]. \\

\begin{figure}[ht]
  \centering
  \begin{minipage}[b]{0.8\textwidth}
    \includegraphics[width=\textwidth]{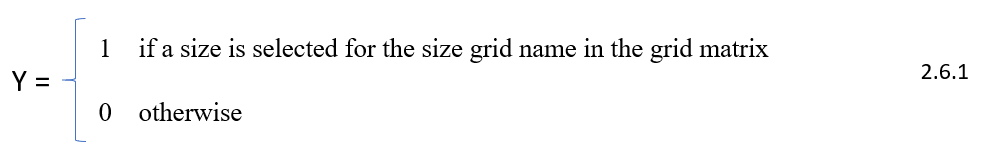}
  \end{minipage}
\end{figure}

In general, dealing with numerical features is common in machine learning. When there are categorical variables, it could be challenging, specially when they have too many levels and are unordered. 5 of the categorical variables in the data have at least 10 levels. The other categorical variables have 2 or 3 levels. Initially, dummy encoding, which convert categorical variables into indicator variables, as one of the encoding methods was used for all categorical variables. This method creates a new column for each level by using 1 and 0’s. Considering the variable with the highest number of levels which is Size Grid Name with 464 levels, there were 464 new attributes for model training. When dummy encoding was applied to all categorical variables, 626 new columns for categorical variables were created. In addition to 75 continuous variables and 50 dummy binary variables, it makes 751 attributes in total. When the dimensions in the data increases, the data can become sparse and it is possible to face with the curse of dimensionality. The necessity of having observations grow exponentially as opposed to the increased number of dimensions in the data [11]. In fact, at least 5 training observations should exist for each dimension in the data [12]. In the study, the data meets this condition by having much bigger sample than the minimum requirement. However, it is still possible that having too many attributes in the data could have negative impact on model performance. Target encoding can perform better than traditional encoding methods for the data with high number of unordered levels in different supervised machine learning models [13]. Therefore, target encoding was also applied to the categorical variables with at least 10 levels. Considering the target variable of the study is a binary and taking values 0 or 1, the categorical features were replaced with a blend of the expected value of the response variable given particular categorical value and the expected value of the target over all the training data by using target encoding [14]. For categorical variable with levels lower than 10, the dummy encoding was remained. By that, 14 categorical variables, 75 continuous variables and 50 dummy variables making 139 in total were features of the model. Comparing to 751 attributes by dummy encoding, the model to be trained had 139 features by using target encoding. Both data sets which were obtained by using only dummy encoding and using both dummy and target encoding were kept separately to be trained in modeling stage. The model performances will be compared and discussed in later stages.\\

Previously it was mentioned that the response variable is a class variable and denoted as 0 and 1 for classes. For the response variable, the data is imbalance where 0’s dominate 1’s (Figure.11). Machine learning models could perform poor with imbalanced data. Accuracy is one of the key performance indicators for classification problems. The standard classification algorithms try to achieve high accuracy by minimizing the misclassification error. With an imbalance dataset, there is relatively little information about the dominated class depending on the imbalance. The model may try to classify all predictions as the dominant class to achieve high accuracy which results in undesired poor classifier as a model. The other problem with imbalance data is most of the classifiers are developed with the assumption of having a balanced dataset [15]. There are over sampling and under sampling methods to overcome this issue. Under sampling for this study was not preferred as no information wanted to be lost by balancing the dominating class to the minority class. For over sampling method, synthetic minority oversampling technique (SMOTE) was chosen. Random sampling methods basically work by taking observations from minor class and duplicate the observations to correct the imbalance between classes. The sampled observation is not a new observation, but a duplicated one that already exist in the minor class. On the other hand, SMOTE oversampling method is synthesizing observations from the minor class. It takes a random data from minor class. Afterwards, k number of the nearest neighbors of the selected data is found and a randomly selected one among the nearest neighbors is picked to create a synthetic observation by forming a line segment in the feature space. It is a type of augmenting the data and could be effective for the model [16]. The models were trained with both oversampled and imbalance data. The model performances will be compared and discussed in later stages.Moreover, parameter tuning with the best performer among the models and cross validation were used in order to see if the model performance could be enhanced.\\

The response variable is a binary class variable so that classification models are considered while evaluating the model selections. From simpler models to more complex models were discussed in this section. Simple models are easier to interpret the results although it brings more assumptions and limitations. In general, more complex models have fewer assumptions and limitations, but more difficult to interpret the coefficients, variables and the results. \\

To begin with, logistic regression is one of the simple classification model approaches to be considered for the study. It assumes the dependent variable to be binary which is the case for our dataset. Also, the assumption of observations being independent of each other is met. They don’t come from repeated measurements or matched data. Little or no multicollinearity among the independent variables is assumed for the logistic regression. It means the attributes of the model should not be highly correlated with each other [17]. However, this assumption is violated. Some independents in our data are highly correlated with each other (See from Figures.83 to Figure.87 in Appendix). Adjusted demand is correlated with Sell-Out where the correlation coefficient is 0.7. This is relatively a high relation to assume they are independent. There are also some other variables which are highly correlated in the data. By the nature of the data for this study, this is expected since all continuous variables are showing the sell performance by using the number of products sold as a basis. However, the three KPI’s have different formulations to represent the sales performance. Finally, a large sample size is required for logistic regression. A guideline stated in one of the recent works is at minimum 10 observations are needed with the least frequent outcome for each independent variable in the model [17]. In our study, the expected probability of the least frequent outcome is 290,067/ 1,225,439 = 0.24. Then minimum sample size required would be (751*10)/0.24 = 31,292, where the sample size of our dataset is much larger. Thus, this assumption was also met for logistic regression. Although one of the assumptions was violated, Logistic Regression was kept as a model to be trained to see how it performs compared to other model approaches. The naïve Bayes classifier is the second model approach for this study, where it basically relies on the estimation of univariate conditional probabilities. Many studies showed that the naïve Bayes classifier is an effective technique for many classification applications with its accurate estimation of univariate conditional probabilities and variable selection. However, it is prone to overfitting [18]. The naïve Bayes classification model requires independence among predictors. Like Logistic Regression, a feature being related to the presence of any other feature violates its assumption. Violation of its independence assumption can harm the performance of the model. Despite this violation, the naïve Bayes classifier was used as one of the model approaches and its performance was discussed. Random Forest Classifier is one of the bagging algorithms where each classifier is built in ensemble using a randomly sampled data and giving an equal vote while classifying the instances. The approach is more robust and can deal with overfitting problems among other ensemble approaches [19]. Random Forest Classifier has some more advantages. It is known as one of the most accurate learning algorithms and still can maintain accuracy when there are a lot of missingness in data. It can handle many features and it is advantageous for this study considering that there are 751 features in our dataset. Random forest algorithm is non-parametric so that it does not require any formal distributions for the variables. Thus, it can handle skewed distributions and ordinal and non-ordinal categorical data where we have in the dataset. On the other hand, it has some limitations as opposed to its many advantages. It can overfit datasets which are particularly noisy which is not the case for our dataset. Moreover, random forests are biased in favor of categorical predictors with more levels. This will be considered in discussion of model results as there are categorical features with different levels in our dataset. The last model approach considered for the study is Extreme Gradient Boosting Classifier (XGBoost Classifier). It is one of the gradient boosting algorithms that was developed recently. Although gradient boosting algorithms are known with their high predictive capabilities, they suffer from very long training times. With the development of XGBoost, it changed the way gradient boosting was done. Data is organized in order to save from lookup times and individual trees are created by multiple cores in XGBoost [20]. Although XGBoost is still using gradient boosting technique at its core, the difference from the simple gradient boosting is that it takes multi-threaded approach in the process of addition of the weak learners. Its proper utilization of the CPU core of the machine makes XGBoost very fast and great performer [21]. In one of the comparative studies, XGBoost was found to be the most accurate classifier among other classifiers including Random Forest [22]. On the other hand, XGBoost can easily overfit if parameters are not tuned properly. To sum up, four different classification algorithms were considered for automating the decision making in the company: Logistic Regression, Naïve Bayes Classifier, Random Forest Classifier and XGBoost Classifier. \\

After having processed the data and obtained the final dataset, it was split into three sets before modeling: train, validation and test. The split was based on the season variable. The size selection decision making is done for an upcoming season by taking previous seasons historical data. Therefore, train set was selected until the Fall-Winter 2019 season and it is 74\% of the sample (Table.1). Validation set represents the Spring-Summer 2020 and has 13\% of the total observations. It will be used to evaluate the model results after training each model. Finally, test set which has 13\% of the total sample is for the Fall-Winter 2020 season of the historical data. After the best performer among the models is selected, the model will be tested with the test set.\\

\begin{table}[!ht]
\begin{center}
\begin{tabular}{ |c|c|c|c|c| }
\hline
\textbf{Season Code} & \textbf{Season} & \textbf{Nb. of records} & \textbf{Split} & \textbf{\%} \\
\hline
{163} & {Fall-Winter 2016} & $55,543$ & & \\
{171} & {Spring-Summer 2017} & $142,757$ & &  \\
{173} & {Fall-Winter 2017} & $167,284$ & &  \\
{181} & {Spring-Summer 2018} & $192,294$ & {Train} & {74\%} \\ 
{183} & {Fall-Winter 2018} & $238,035$ & & \\ 
{191} & {Spring-Summer 2019} & $214,502$ & & \\ 
{193} & {Fall-Winter 2019} & $215,024$ & & \\ 
\hline
{201} & {Spring-Summer 2020} & $215,993$ & {Validation} & {13\%} \\
\hline
{203} & {Fall-Winter 2020} & $217,586$ & {Test} & {13\%} \\
\hline
\end{tabular}
\end{center}
\caption{Data split into train, validation and test sets}
\end{table}

4 classifiers will be trained initially with the dataset where all categorical variables were encoded by dummy encoding (Model I in Figure.22). Afterwards, all four classifiers will be trained with the dataset where the categorical variables with at least 10 levels were encoded by target encoding and the categorical variables with lower than 10 levels were encoded by dummy encoding (Model II in Figure.22). In both Model I and II steps, the models will be trained without oversampling. After applying SMOTE oversampling to balance the classes in the response variable, all 4 classifiers will be trained with the dataset where target encoding was applied before (Model III in Figure.22). Afterwards, the best performer among the classifiers will be selected and parameter tuning with cross validation will be applied for the selected classifier (Model IV in Figure.22). \\

\bxfigure{Model training steps with performance boosting techniques. I, II, III and IV represents the steps and at each step, except IV, all 4 classifiers were trained. Only XGBoost Classifier was trained at Step IV. }{\includegraphics[scale=1]{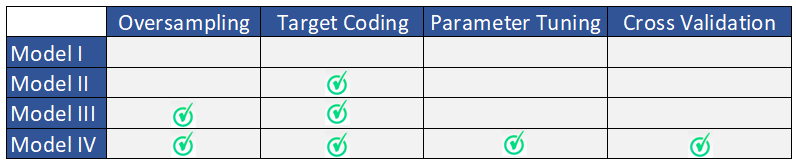}}

As it is a binary classification approach to the problem, the area under the curve (AUC) will be one of the measurements showing the ability of a classifier to distinguish between classes. Accuracy showing the fraction of the correct prediction to all predictions is the second performance metric to be measured while evaluating the model results. Precision and recall are the other two metrics to be measure during the model evaluation. Precision showing the fraction of true positives among all positives is needed as the cost of false positives could be high. If a size is selected poorly, many products will be produced from that size which can result in thousands of products as excess stock in inventory when they are not sold. This will increase production and inventory costs with no revenue in return. In fact, if the poorly selected size belongs to a seasonal product, it will be even more costly as seasonal products will not be in the market after a season ends. Recall is the metric showing the fraction of true positives among true positives and false negatives. For the study, recall is also a significant metric as the cost of false negatives could be high. If a size which has a selling potential in big volumes is not selected in the grid matrix, no products from that size will be produced. It will cause missing sell opportunities by not producing a size corresponding to the actual customer need.\\

\addcontentsline{toc}{section}{2.7 Results}
\subsection*{2.7 Results}
The model results will be discussed in this section. Initially, all 4 classifiers -Logistic Regression, Naïve Bayes Classifier, Random Forest Classifier and XGBoost Classifier- were trained without oversampling, target encoding or cross validation. Before starting the model steps defined in Figure.22, the models were trained without handling the missing data to observe how model perform before and after missing data handling. The results from the models can be seen in Table.2. Random Forest gave better results than others except its precision score which is worse than XGBoost. ROC curves were constructed and can be found in Figure.23. ROC by Random Forest Classifier presents the best curve to obtain the highest True Positive Rate and the lowest False Positive Rate. \\

\bxfigure{ROC curves for the classifiers: LR:Logistic Regression, NB: Naïve Bayes Classifier, RF: Random Forest Classifier. NB gives relatively poor ROC curve compared to others while RF and XGBoost curves are close to each other and RF gives slightly a better curve}{\includegraphics[scale=0.8]{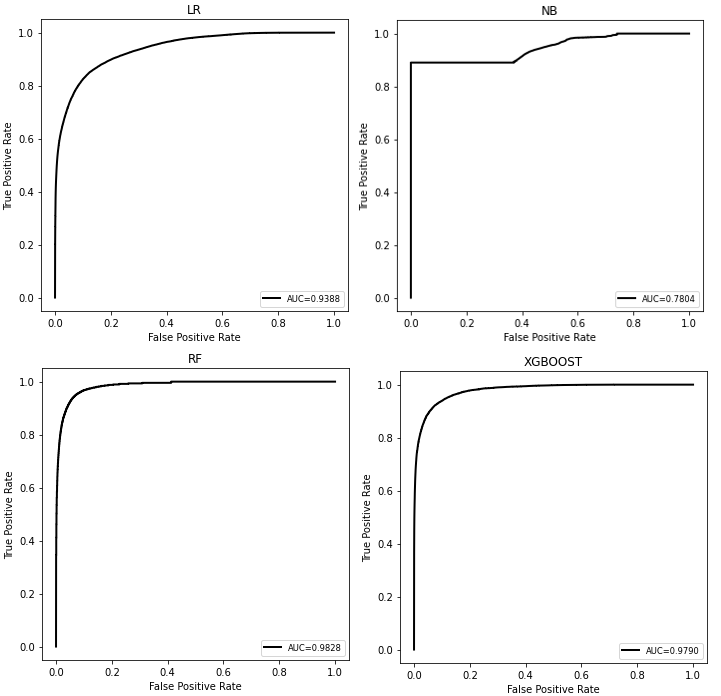}}

\begin{table}[!ht]
\begin{center}
\begin{tabular}{ |c|c|c|c|c| }
\hline
\textbf{Metrics} & \textbf{Logistic Regression} & \textbf{Naïve Bayes} & \textbf{Random Forest} & \textbf{XGBoost} \\
\hline
{AUC} & $0.9388$ & $0.7804$ & $0.9828$ & $0.9790$ \\
\hline
{Accuracy} & $0.9060$ & $0.6083$ & $0.9499$ & $0.9441$ \\
\hline
{Precision} & $0.8123$ & $0.3392$ & $0.8782$ & $0.8910$ \\
\hline
{Recall} & $0.7076$ & $0.9487$ & $0.8787$ & $0.8306$ \\
\hline
\end{tabular}
\end{center}
\caption{Performance metrics of the models before missing data handling}
\end{table}

After handling the missing data, obtained results are presented in Table.3. It was seen that all performance metrics were improved for all 4 classifiers. The differences between two models are remarkable, specially for Logistic Regression. The improvements were 0.0606 for AUC, 0.0871 for accuracy, 0.1705 for precision and 0.2762 for recall. Without handling the missing data, the precision and recall metrics were poorer for all classifiers. Looking the accuracy metrics, Random Forest and XGBoost classifiers were impacted less than Logistic Regression and Naïve Bayes. This implies that these models were more robust to missing data. The rest of the discussions will be made based on the models which were trained after having handled the missingness.\\

Naïve Bayes had the lowest AUC score, 0.8776 (Table.3). Logistic Regression follows Naïve Bayes with AUC 0.9995. Logistic Regression has more predictive ability than Naïve Bayes in discriminating the classes. Random Forest and XGBoost have AUC scores close to each other, respectively 0.9997 and 0.9999. Random Forest and XGBoost almost perfectly discriminate between classes with high AUC scores where the score of XGBoost is slightly higher than Random Forest.\\

In terms of accuracy, XGBoost is the best model while predicting the true positives and negatives with 0.9964 accuracy score. Random Forest is almost as good as XGBoost in terms of accuracy with 0.9957 accuracy score. Logistic Regression also performs well in correctly predicting classes (with accuracy score 0.9931), but worse than XGBoost and Random Forest. On the other hand, Naïve Bayes results were relatively poor in predicting the correct classes with 0.7277 accuracy score. With the precision score 0.9828, Logistic Regression classifier can predict most of the positives correctly. Recall score for the same classifier is 0.9838. Naïve Bayes classifier can predict only 43\% of the positives correctly. This is relatively a poor score for a classifier. However, the recall score of Naïve Bayes is 0.9959 which is better than the recall score of Logistic Regression by 1\%. Both Random Forest Classifier and XGBoost Classifier are very good performers in terms of precision and recall and have very close precision and recall scores to each other. The precision scores for Random Forest and XGBoost are respectively 0.9811 and 0.9864. The percentages of the total relevant results correctly classified by Random Forest and XGBoost classifiers are high and close to each other in magnitude, respectively 0.9982 and 0.9964 (Table.3). XGBoost is the best performer among all classifiers with the highest performance metrics mostly.\\

\begin{table}[!ht]
\begin{center}
\begin{tabular}{ |c|c|c|c|c| }
\hline
\textbf{Metrics} & \textbf{Logistic Regression} & \textbf{Naïve Bayes} & \textbf{Random Forest} & \textbf{XGBoost} \\
\hline
{AUC} & $0.9995$ & $0.8776$ & $0.9997$ & $0.9999$ \\
\hline
{Accuracy} & $0.9931$ & $0.7277$ & $0.9957$ & $0.9964$ \\
\hline
{Precision} & $0.9828$ & $0.4307$ & $0.9811$ & $0.9864$ \\
\hline
{Recall} & $0.9838$ & $0.9959$ & $0.9982$ & $0.9964$ \\
\hline
\end{tabular}
\end{center}
\caption{Performance metrics of the models}
\end{table}

In the second step, the models were trained with the data where target encoding was applied to the categorical variables with at least 10 levels. The performance metrics of Logistic Regression, Random forest Classifier and XGBoost Classifier were impacted with very slight changes while the impact for Naïve Bayes was larger on the metrics. Recall score for Logistic regression dropped by 0.0002, while the AUC, accuracy and precision scores stayed the same (Table.4). Naïve Bayes classifier was able to discriminate the classes better with target encoding with 0.9797 AUC score. Its accuracy became 0.9597 (higher than the first model by 0.1) and the precision score almost doubled to 0.8436. Naïve Bayes was able to classify positives much better with fewer features although it has a capability of learning better with more features in general. AUC, accuracy and precision scores were better with target encoding for Random Forest, respectively 0.9998, 0.9960 and 0.9832 (Table.4). However, its recall score decreased by 0.0004. Among all classifiers, XGBoost was the model which was impacted the least. Its AUC and accuracy scores did not change. The precision score decreased by 0.0001 and recall score decreased by 0.0004.\\

\begin{table}[!ht]
\begin{center}
\begin{tabular}{ |c|c|c|c|c| }
\hline
\textbf{Metrics} & \textbf{Logistic Regression} & \textbf{Naïve Bayes} & \textbf{Random Forest} & \textbf{XGBoost} \\
\hline
{AUC} & $0.9995$ & $0.9797$ & $0.9998$ & $0.9999$ \\
\hline
{Accuracy} & $0.9931$ & $0.9597$ & $0.9960$ & $0.9964$ \\
\hline
{Precision} & $0.9828$ & $0.8436$ & $0.9832$ & $0.9865$ \\
\hline
{Recall} & $0.9836$ & $0.9875$ & $0.9978$ & $0.9960$ \\
\hline
\end{tabular}
\end{center}
\caption{Performance metrics of the models after target encoding}
\end{table}

In the third step, oversampling was applied to the data since the imbalance between classes in the data could result in poorer performances for the models. This is normally caused by that if a model is trained with very few positives, it does not learn enough to classify them and negatives dominating positives makes the model tend to predict them as negatives to increase the accuracy. In the data of this study, there were 282,234 positives which could be enough for a model to be trained. Yet, it was still possible to enhance the model results although they were satisfying specially for XGBoost. With oversampling, very slight changes were observed in the performance metrics of the models (Table.5). Recall of Naïve Bayes increased by 0.0008 while the precision score went down by 0.0013. AUC score for XGBoost stayed the same. Unlike Random Forest, XGBoost is also known as offering good performance on classification problems with severe imbalance. A decrease by 0.0002 and 0.0001 was observed in the accuracy scores for them in order. Similarly, precision dropped by 0.0014 and 0.0003 for Random Forest and XGBoost respectively while the recall scores went up by 0.0005 and 0.0001. The biggest impact on the models was for Logistic Regression. While the precision decreased by 0.0039, the accuracy and recall metrics were enhanced by 0.0004 and 0.0065 respectively. \\

\begin{table}[!ht]
\begin{center}
\begin{tabular}{ |c|c|c|c|c| }
\hline
\textbf{Metrics} & \textbf{Logistic Regression} & \textbf{Naïve Bayes} & \textbf{Random Forest} & \textbf{XGBoost} \\
\hline
{AUC} & $0.9995$ & $0.9798$ & $0.9997$ & $0.9999$ \\
\hline
{Accuracy} & $0.9935$ & $0.9594$ & $0.9958$ & $0.9963$ \\
\hline
{Precision} & $0.9789$ & $0.8423$ & $0.9818$ & $0.9862$ \\
\hline
{Recall} & $0.9901$ & $0.9883$ & $0.9983$ & $0.9961$ \\
\hline
\end{tabular}
\end{center}
\caption{Performance metrics of the models after target encoding and oversampling}
\end{table}

Although it may seem that the differences between the metrics of 4 classifiers are not too big, the impact on sales volumes could be much more. A couple of misclassification errors could correspond to a loss of thousands of products and millions of Euros for the company. Therefore, the impact was measured for the errors (Figure.24). The impact showed Logistic Regression and Naive Bayes could cause more loss in sales than Random Forest and XGBoost. In fact, Naive Bayes was performing poorer than Logistic Regression. However, Logistic Regression has the largest loss in sales for the company (Figure.24). This shows Logistic Regression was less capable of selecting sizes for volume driver products compared to other classifiers. The difference between Random Forest and XGBoost was much less in terms of both volumes and revenue loss. While Random Forest was causing less loss in volumes, XGBoost was causing less money loss. The differences are on season level and both models perform close to each other, where XGBoost is a bit better when one focuses on the financial impact. XGBoost was selected as the best model among the other models after having analyzed the performance metrics and the impact of the models on sales. In fact, both Random Forest and XGBoost classifiers performed well. XGBoost was slightly better than Random Forest. In addition, Random Forest was more time costly than XGBoost in terms of training. \\

\bxfigure{Loss of sales in volumes and revenue (EUR) explains the impact of the models for the company}{\includegraphics[scale=1]{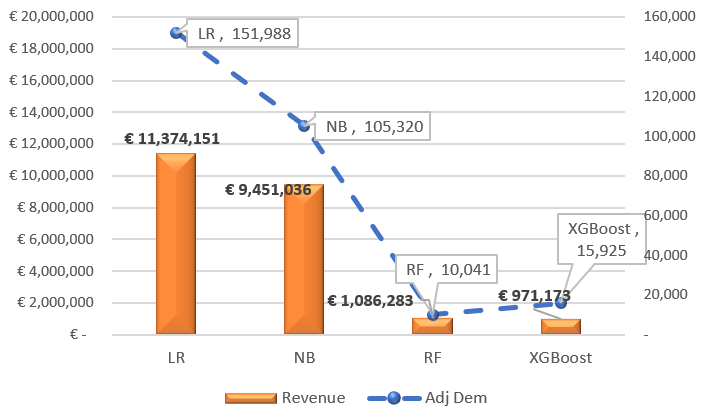}}

As a further step, parameter tuning was done for XGBoost. GridSearchCV was used for parameter tuning. It uses exhaustive search method over specified parameter values for an estimator [23]. The given parameters to be searched for tuning were maximum depth parameter, number of estimators and learning rate. Initially, integers in range [2, 10) for maximum depth, numbers between [60, 220) increasing by 40 for number of estimators and three rate values 0.05, 0.01 and 0.1 for learning rates were given. Additionally, 3 folds were given as a parameter for cross validation and accuracy was selected for scoring. 0.05 for learning rate, 4 for maximum depth and 60 for number of estimators were selected as the best parameters. However, the minimum number in the range that was given as parameters for number of estimators was 60. It was still possible to go beyond the minimum range. For this reason, it was rerun with the range [10, 60) increasing by 10 for n estimators. 40 for the parameter was selected as the best parameters this time. After having decided the best parameters for XGBoost classifier, learning curve was plotted in order to check if there is overfitting or underfitting for the model before running the trained model with the test data. Around sample size 1000, the convergence between training and validation was obtained. Additionally, validation accuracy was smaller than training accuracy and the difference between them is relatively acceptable (Figure.25). The result was that there is no underfitting or overfitting for the selected model.\\

\bxfigure{Learning curve for XGBoost}{\includegraphics[scale=1.5]{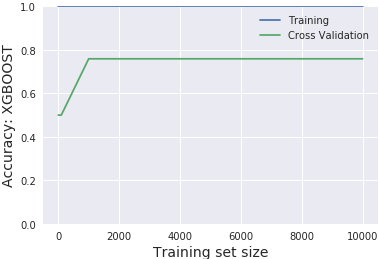}}

There were 173,823 negatives and 48,168 positives in the test set. After having trained the model with the best parameters, the predictions were made with the test set. The AUC score was 0.9998 for the test set. 0.9968 accuracy was obtained where 173,126 + 48,156 = 221,282 number of the predictions were classified correctly (Figure.26). 697 sizes were not selected by the model where the selection was made for them in real (false negatives). 12 sizes were selected by the model when they were not selected in real (false positives). False positive rate is 12 / (12 + 173,126) = 0.01\%. Sensitivity (recall) was 48,156 / (48,156 + 697) = 0.9857 and specificity was 173,126 / (173,126 + 12) = 0.9999. In total, 697+12 = 709 predictions were made incorrectly. Misclassification rate was 709 / (173,126 + 48,156) = 0.0032. \\

\bxfigure{Confusion Matrix}{\includegraphics[scale=1.2]{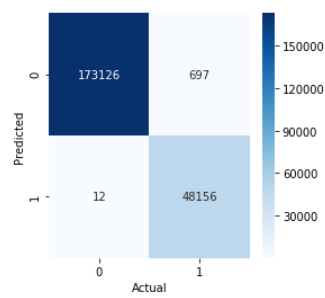}}

One aspect of this study is that size selection is made for a certain planning group and a group of products. Therefore, 1 misclassified prediction could mean thousands of products. In the test set, 697 predictions were misclassified where they were actually selected by the planners for Fall-Winter 2020. The demand generated for this set is 20,629 products in units. Its revenue impact is EUR 1,421,158.93. This means if the model had been used for size selection of Fall-Winter 2020, the demand with 20,629 of products would have been missed because of the classification error. The false positives were 12 for the test set. However, since the model selected these sizes and they were not selected by the planners, it was not possible to measure the impact of the false positives. 

\addcontentsline{toc}{section}{3. DISCUSSION}
\section*{3. DISCUSSION}

In this section, the results of the experiments were summarized and how the experiments helped for answering the research questions were explained. Subsequently, main contributions and future work were explained.

\addcontentsline{toc}{section}{3.1 Conclusion}
\subsection*{3.1 Conclusion}
The goal of this study was to automate size grid building and management process by machine learning techniques for an apparel and fashion company. The process is manually managed, where each stage is managed by different stakeholders in the company. Size grids are built by sizing team on size grid name level, which represent the group of certain products that will be in the market for the planned season. After the size grid are built and put in the internal systems, they are made available to the front end of users. Planners as retail and wholesale make their size selections for each planning group based on the company strategies and customer requests. Both size grid building and size selection decisions are made based on some key performance indicators that are representative of sales performance for the company. The three stages in the process -size grid building, data preparation{\&} system feed, size selection- require manual workload.\\

Some challenges were faced in the the study. Firstly, data for some KPI's was not available for all stores. Missing observations were needed to be handled. Dummy binary variables were added into the dataset where 1's indicated the missing data and 0's were for the rest. Secondly, size grid names were not stationary and structured in the history, which was important to build a machine learning model. The products under the size grid names were mapped so that a product can be grouped under only one size grid name in the history. Finally, all data sets were available for selected sizes. Candidate sizes which were considered during the size grid building did not exist. Therefore, a data set was needed to be created. However, sizes in the grid matrix of a size grid name did not have a unique format neither in different size grid names nor in the same size grid. Manually a data set was created to obtain candidate sizes for each size grid name.\\ 

This study was designed as a classification problem, where the target variable is whether to select a size in the grid matrix or not. The data set was splited into train, validation and test based on the season. Four classifiers were considered as model approaches to the problem: Logistic Regression, Naive Bayes Classifier, Random Forest Classifier and XGBoost Classifier.  Independence assumption among the attributes was violated for both Logistic Regression and Naive Bayes as there were high correlations between some independents. Area under ROC curve, accuracy, precision and recall were selected as the metrics for the evaluation of the models.
There was imbalance between the classes in the data. Additionally, some categorical variables having too many levels resulted in many attributes after encoding them. Oversampling, target encoding, cross validation and parameter tuning methods were applied in order to enhance the model performances.\\

Logistic Regression, Random Forest and XGBoost were able to give satisfying results in the first step. Although Naive Bayes was not as good as the other models, it did not give poor results except precision. After target encoding, Naive Bayes metrics were improved. Oversampling did not have a major impact on the models. It means the models were able to predict the classes with imbalance data. It can be said that dominated class having around 200k observations were enough for them to be trained to make accurate predictions. XGBoost was selected as the best model with slightly better metrics than Random Forrest classifier. Test set was representative for Fall-Winter 2020. Trained model with XGBoost after parameter tuning was tested on the set. Similar to the validation set, the results obtained with the test test were also satisfying. \\

\addcontentsline{toc}{section}{3.2 Contributions}
\subsection*{3.2 Contributions}
Fashion and apparel industry suffers from quickly changing trends and very short seasons as opposed to long lead times of production. This makes season planning for fashion and apparel companies harder every year. A more automated process ,which is one of the components in season planning, was created with the developed model. A model which will help sizing planners to build size grids and planners to make size selections was built by using machine learning techniques. After it is put into practice, the manual workload will decrease for these teams. The size grid building and management process is performed for each season. The company is planning to have more seasons in a year, which means the model will create even bigger difference by taking over the workload that will be created in the future. Many studies focusing on machine learning for fashion and apparel industry are designed to make predictions based on products. Because of the variability and robustness concerns, size of a product is out of scope in general. This study is focusing on product sizes and uses machine learning techniques to make predictions by putting the size in the center of scope. Additionally, XGBoost as one of the recently developed technique was applied to the clothing data. It proved that it can be also helpful in solving the problems of fashion and apparel companies.

\addcontentsline{toc}{section}{3.3 Future Work}
\subsection*{3.3 Future Work}
The target variable of the model in this study was based on the human decisions made in the history. However, it was possible to make better decision making in size selections. An alternative model can be conducted in the future where its target variable is not dependent on the decisions made in the past but more on the optimal decisions that could have been made.  Moreover, size grid names are actually classes of grouped products where the classes were based on some characteristics of the products. Yet, they are grouped manually. This also causes them being changed and updated frequently. As a further research, a machine learning classification method can be used in order to come up with more representative groups for products. By that, the size grid names will be structured by scientific methods and can be more stationary through the seasons. In fact, the segmentation approach can be helpful also to involve new size grid names in the model, which are currently managed separately. \\

\newpage

\newpage

\section*{References}
[1] Vashishtha R.K., Burman V., Kumar R. (2020). Product age-based demand forecast model for fashion retail, California-USA.\\

[2] Clothing Sizes, Wikipedia: https://en.wikipedia.org/wiki/Clothing\_sizes \\

[3] Varalica J. (2018). ISO Standards and European Norms for Size Designation of Clothes. Journal of Textile Science {\&} Engineering. Department of Clothing Science and Engineering Technology, University of Zagreb, Croatia \\

[4] Kuhn M.G., (2003). BodyDim: Body dimension pictograms for size designation of clothes generated with Meta Post. P 08-13 \\

[5] Nenni M. E., Giustiniano L. and Pirolo L. (2013). Demand Forecasting in the Fashion Industry: A Review. International Journal of Engineering Business Management Special Issue on Innovations in Fashion Industry \\

[6] Ferreira K. J., Simchi-levi D. (2015). Analytics for an Online Retailer: Demand Forecasting and Price Optimization. Manufacturing {\&} Service Operations Management. \\

[7] Flegal KM (2007) Waist circumference of healthy men and women in the United States. International Journal of Obesity 31, 1134-1139.\\

[8] Hastie T., Tibshirani R., Friedman J. (2008) The Elements of Statistical Learning. Springer Series in Statistics. Stanford, California 398-400 \\

[9] Scikit learn API. https://scikit-learn.org/stable/modules/generated/sklearn.preprocessing.\ RobustScaler.html \\

[10] Scikit learn API: https://scikit-learn.org/stable/modules/generated/sklearn.preprocessing. \ StandardScaler.html \\

[11] Curse of dimensionality https://en.wikipedia.org/wiki/Curse\_of\_dimensionality \\

[12] Theodoridis S., Koutroumbas K. (20008) Pattern Recognition. USA \\

[13] Pargent F., Pfisterer F., Thomas J., Bischl B. (2021) Regularized target encoding outperforms traditional methods in supervised machine learning with high cardinality features. Computational Statistics and Data Analysis. Germany \\

[14] Scikit learn https://contrib.scikit-learn.org/category\_encoders/targetencoder.html \\

[15] Ali A., Shamsuddin S.M., Ralescu A.L. (2013) Classification with class imbalance problem: a review. International J. Advance Soft Compu. App, Vol. 5, No.3, Malaysia \\

[16] Chawla N.V., Bowyer K.W., Hall L.O., Kegelmeyer W.P. (2002) SMOTE: Synthetic Minority Over-sampling Technique. Journal of Artificial Intelligence Research, Volume 16, p. 321-357 \\

[17] Schreiber-Gregory D., Bader K. (2018) Logistic and Linear Regression Assumptions: Violation Recognition and Control. SESUG p.247 \\

[18] Boulle M. (2007) Compression-Based Averaging of Selective Naive Bayes Classifiers. Journal of Machine Learning Research. p.1659-1685. France.\\

[19] Fawagreh K., Gaber M.M. (2014) Random forests: from early developments to recent advancements. Systems Science and Control Engineering. p.602-609 \\

[20] Chen T., Guestrin C. (2016) XGBoost : A scalable tree boosting system \\

[21] Ramraj S., Uzir N., Raman S., Banerjee S. (2016) Experimenting XGBoost Algorithm for Prediction and Classifi cation of Different Datasets. International Journal of Control Theory and Applications. V.9 N.40 \\

[22] Bentejac C., Csorgo A., Martinez-Munoz G (2019) A Comparative Analysis of XGBoost. \\

[23] GridSearchCv in Scikit Learn: https://scikit-learn.org/stable/modules/generated/sklearn. \ model\_selection.GridSearchCV.html \\

\newpage
\pagenumbering{gobble}
\section*{Appendix}

\vspace{2mm}
\begin{figure}[ht]
  \centering
  \begin{minipage}[b]{1\textwidth}
    \includegraphics[width=\textwidth]{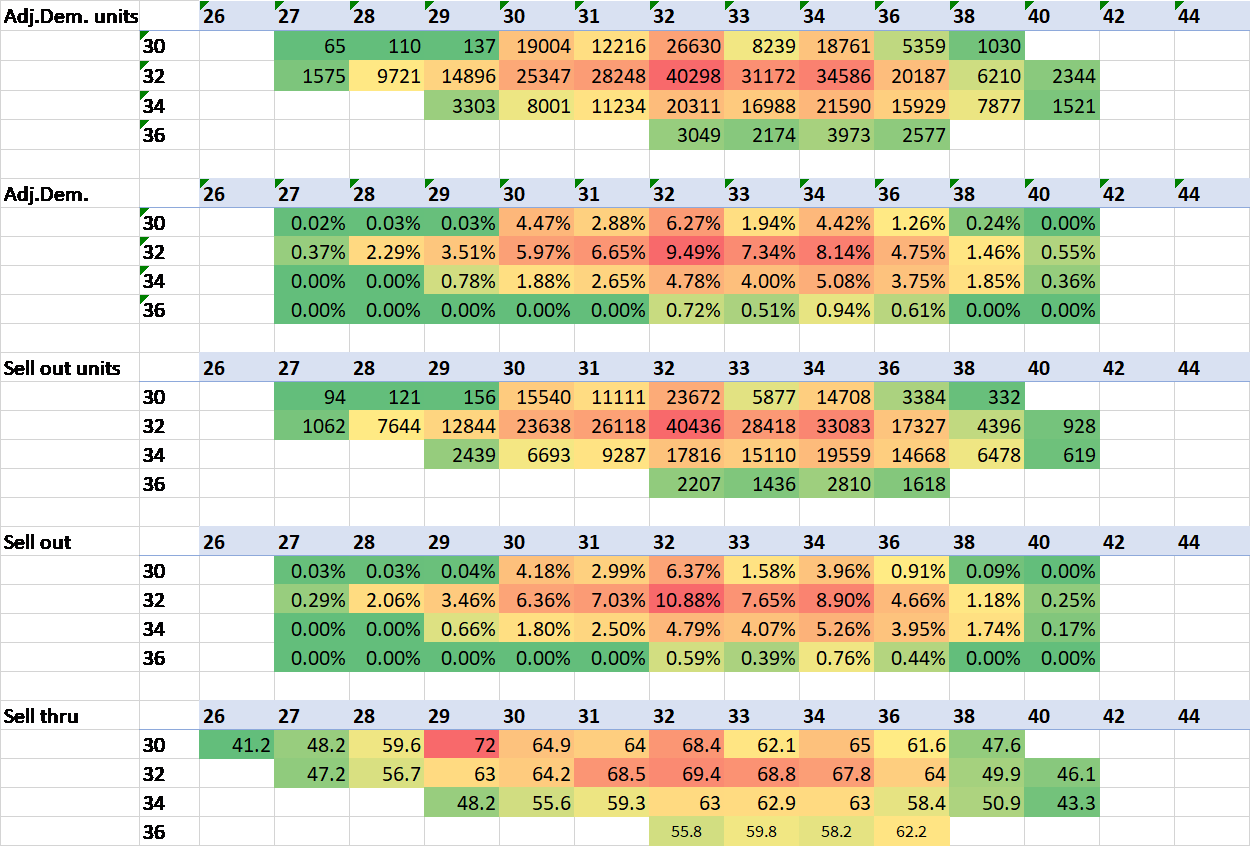}
    \caption{An example of a size grid and its KPI performance}
  \end{minipage}
\end{figure}

\vspace{2mm}
\begin{figure}[ht]
  \centering
  \begin{minipage}[b]{1\textwidth}
    \includegraphics[width=\textwidth]{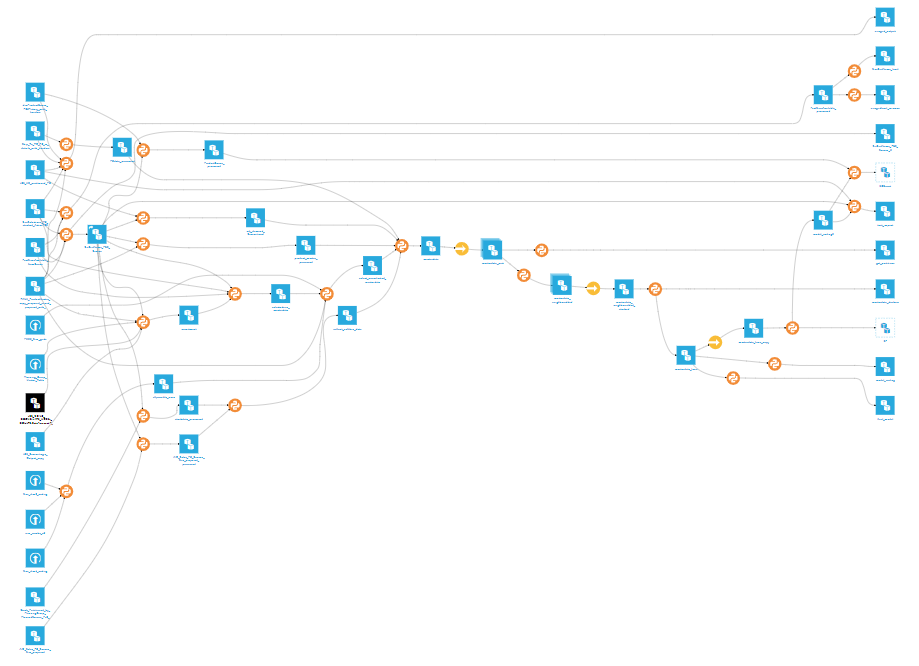}
    \caption{Size Grid Building and Management Project Flow in Dataiku:\textit{Blue boxes} show the data tables, \textit{Orange circle symbols} are python recipes}
  \end{minipage}
\end{figure}

\begin{table}[!ht]
\begin{center}
\begin{tabular}{|p{4cm}|p{9cm}|p{3cm}|}
\hline
\textbf{Dataset Name} & \textbf{Explanation} & \textbf{Nb. of records} \\
\hline
{Item Plan Forecast} & {Which product is sold to which Planning Group on which month} & $10,987,588$ \\
\hline
{Season Logic} & {Which product is sold in which season and month} & $790,979$ \\
\hline
{Fact Sizes Available Conso} & {Region level sizes} & $340,438$ \\
\hline
{Size Selections} & {Planning group level size selections made in SGMT tool} & $6,286,148$ \\
\hline
{Product Master} & {Product related variables (Category text, gender text, target consumer, textile related variables, etc.)} & $618,045$ \\
\hline
{Product Conso} & {Product related market variables (FoF, LoF, PC5 History, Lifecycle, Status, etc.  )} & $97,423$ \\
\hline
{Planning Group} & {Which PG belongs to which channel, affiliate} & $49$ \\
\hline
{Adjusted Demand Historical} & {Historical demand data} & $8,176,133$ \\
\hline
{Stock data} & {Stock left in inventory} & $99,949,863$ \\
\hline
{Sell out Data} & {Products sold in the stores} & $2,657,773$ \\
\hline
{Planning Group Mapping} & {Merged planning groups and their mapping} & $336$ \\
\hline
\end{tabular}
\end{center}
\caption{Datasets and explanations}
\end{table}

\vspace{2mm}
\begin{figure}[ht]
  \centering
  \begin{minipage}[b]{0.7\textwidth}
    \includegraphics[width=\textwidth]{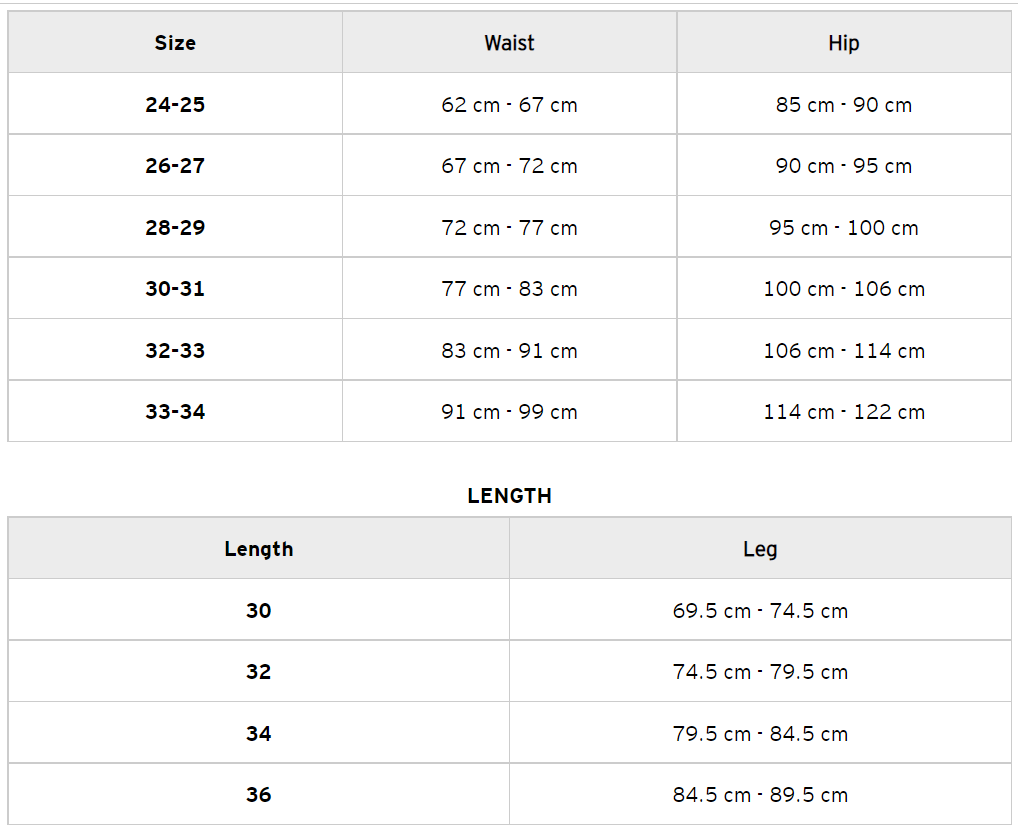}
    \caption{Size charts for female bottoms}
  \end{minipage}
\end{figure}

\vspace{2mm}
\begin{figure}[ht]
  \centering
  \begin{minipage}[b]{0.7\textwidth}
    \includegraphics[width=\textwidth]{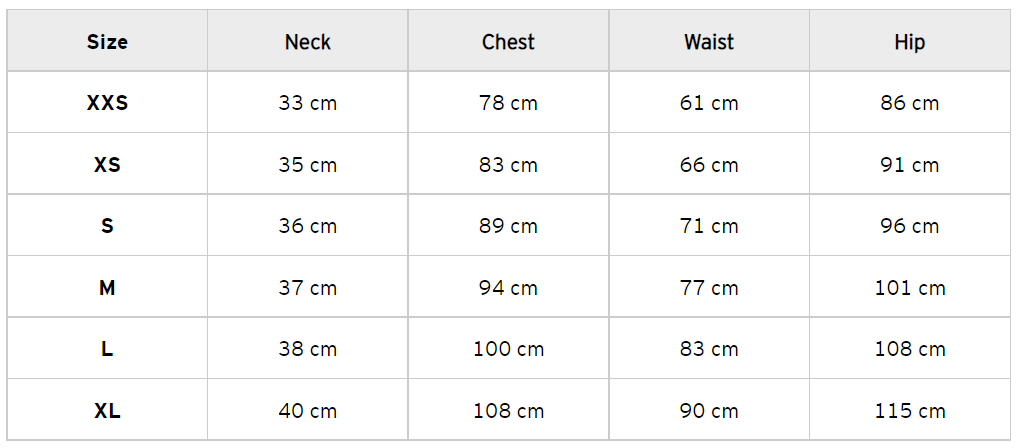}
    \caption{Size chart for female tops}
  \end{minipage}
\end{figure}

\vspace{6mm}
\begin{figure}[p]
  \centering
  \begin{minipage}[b]{0.7\textwidth}
    \includegraphics[width=\textwidth]{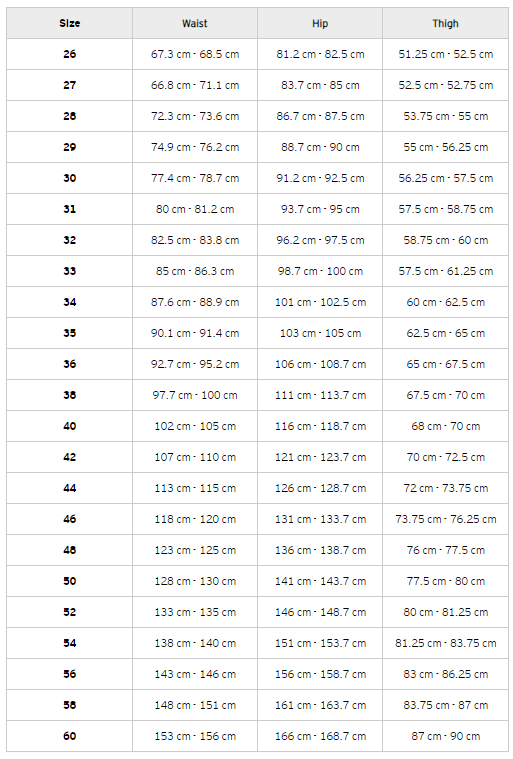}
  \end{minipage}
  \hfill
  \begin{minipage}[b]{0.7\textwidth}
    \includegraphics[width=\textwidth]{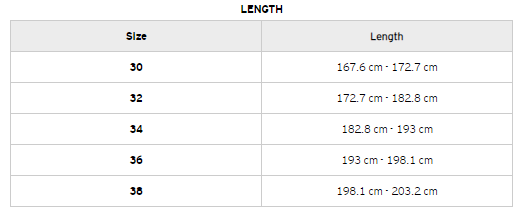}
    \caption{Size charts for male bottoms}
  \end{minipage}
\end{figure}

\vspace{2mm}
\begin{figure}[ht]
  \centering
  \begin{minipage}[b]{0.7\textwidth}
    \includegraphics[width=\textwidth]{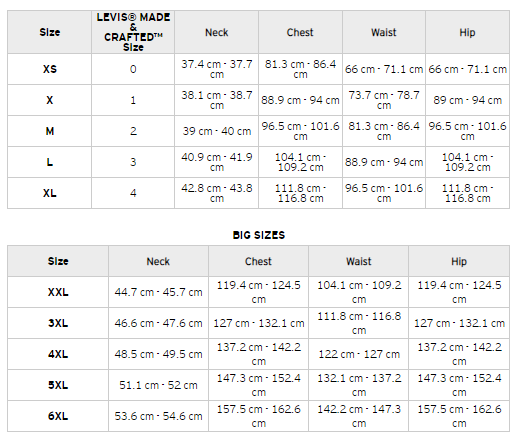}
    \caption{Size chart for male tops}
  \end{minipage}
\end{figure}

\vspace{6mm}
\begin{figure}[ht]
  \centering
  \begin{minipage}[b]{0.45\textwidth}
    \includegraphics[width=\textwidth]{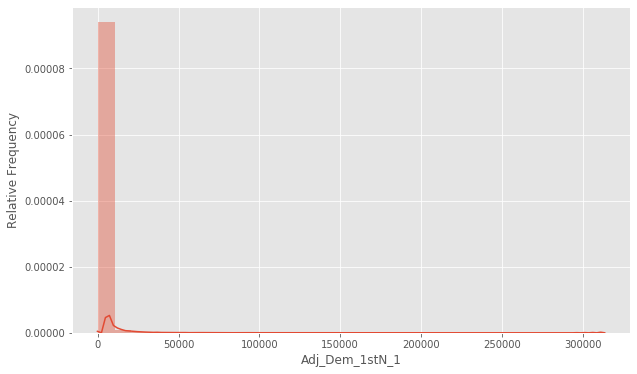}
    \caption{Distribution of Adjusted Demand Neighbor Size-1 in 1.circle without zero values}
  \end{minipage}
  \hfill
  \begin{minipage}[b]{0.45\textwidth}
    \includegraphics[width=\textwidth]{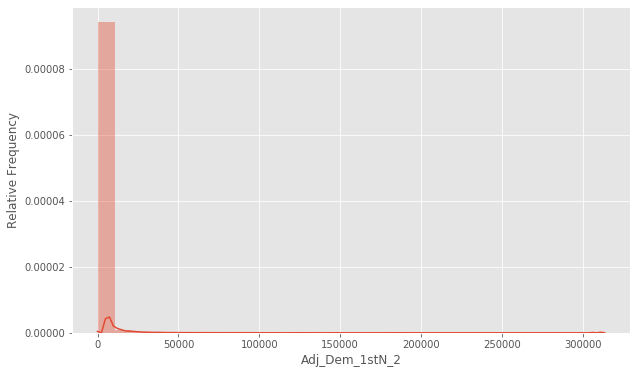}
    \caption{Distribution of Adjusted Demand Neighbor Size-2 in 1.circle without zero values}
  \end{minipage}
\end{figure}

\vspace{6mm}
\begin{figure}[ht]
  \centering
  \begin{minipage}[b]{0.45\textwidth}
    \includegraphics[width=\textwidth]{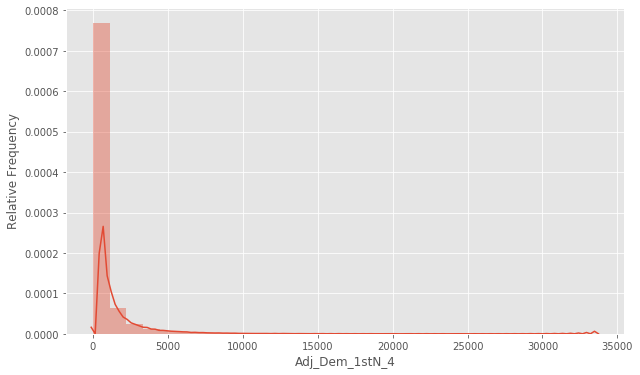}
    \caption{Distribution of Adjusted Demand Neighbor Size-4 in 1.circle without zero values}
  \end{minipage}
  \hfill
  \begin{minipage}[b]{0.45\textwidth}
    \includegraphics[width=\textwidth]{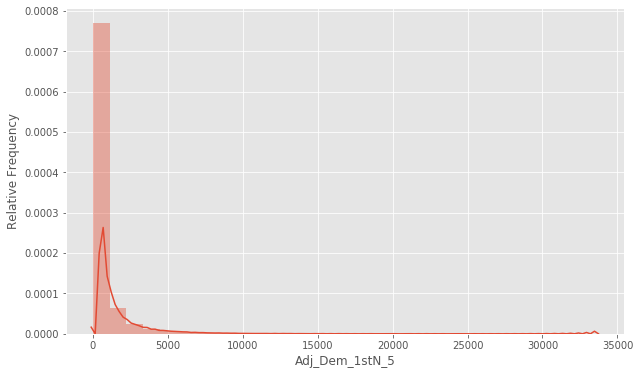}
    \caption{Distribution of Adjusted Demand Neighbor Size-5 in 1.circle without zero values}
  \end{minipage}
\end{figure}

\vspace{6mm}
\begin{figure}[ht]
  \centering
  \begin{minipage}[b]{0.45\textwidth}
    \includegraphics[width=\textwidth]{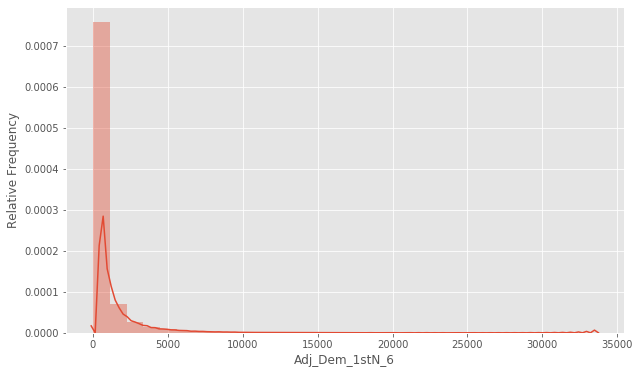}
    \caption{Distribution of Adjusted Demand Neighbor Size-6 in 1.circle without zero values}
  \end{minipage}
  \hfill
  \begin{minipage}[b]{0.45\textwidth}
    \includegraphics[width=\textwidth]{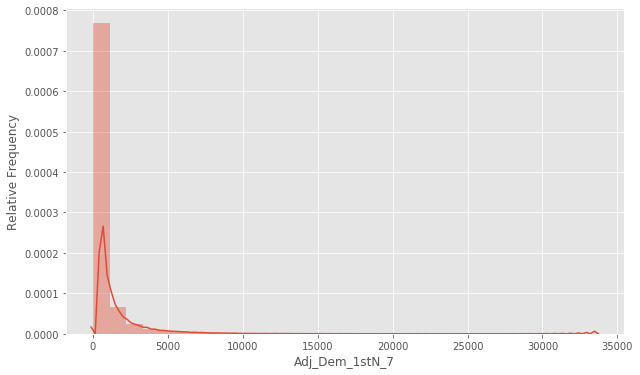}
    \caption{Distribution of Adjusted Demand Neighbor Size-7 in 1.circle without zero values}
  \end{minipage}
\end{figure}

\vspace{6mm}
\begin{figure}[ht]
  \centering
  \begin{minipage}[b]{0.45\textwidth}
    \includegraphics[width=\textwidth]{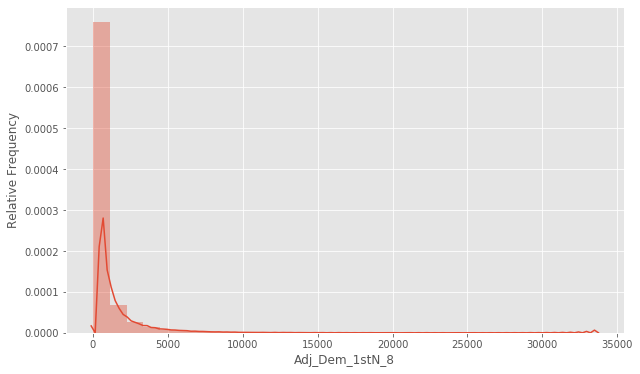}
    \caption{Distribution of Adjusted Demand Neighbor Size-8 in 1.circle without zero values}
  \end{minipage}
  \hfill
  \begin{minipage}[b]{0.45\textwidth}
    \includegraphics[width=\textwidth]{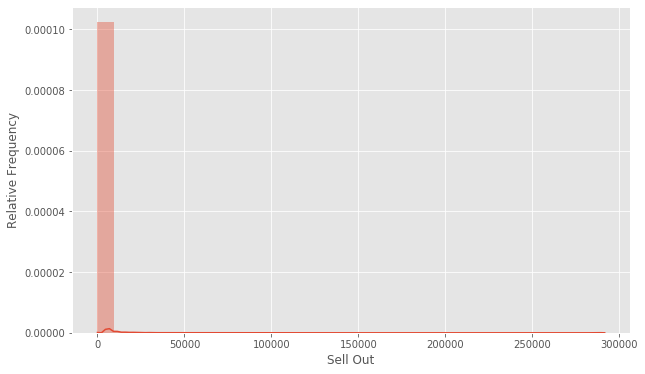}
    \caption{Distribution of Sell Out without zero values}
  \end{minipage}
\end{figure}

\vspace{6mm}
\begin{figure}[ht]
  \centering
  \begin{minipage}[b]{0.45\textwidth}
    \includegraphics[width=\textwidth]{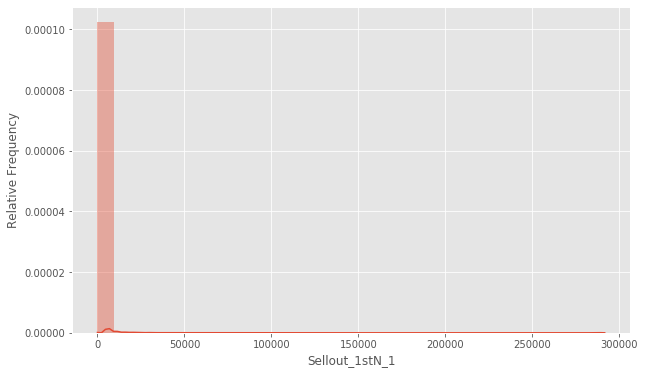}
    \caption{Distribution of Sell Out Neighbor Size-1 in 1.circle without zero values}
  \end{minipage}
  \hfill
  \begin{minipage}[b]{0.45\textwidth}
    \includegraphics[width=\textwidth]{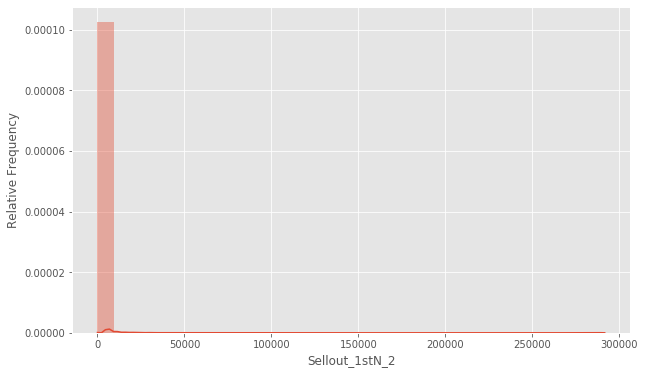}
    \caption{Distribution of Sell Out Neighbor Size-2 in 1.circle without zero values}
  \end{minipage}
\end{figure}

\vspace{6mm}
\begin{figure}[ht]
  \centering
  \begin{minipage}[b]{0.45\textwidth}
    \includegraphics[width=\textwidth]{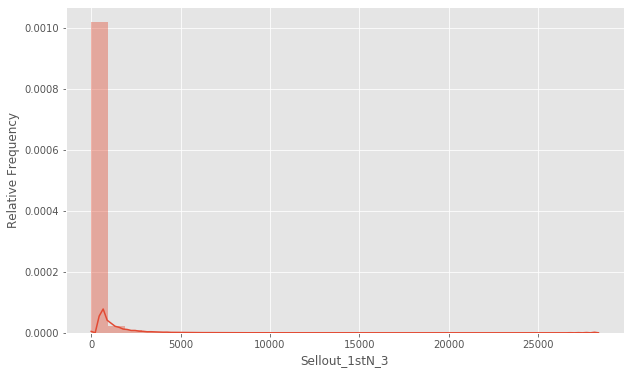}
    \caption{Distribution of Sell Out Neighbor Size-3 in 1.circle without zero values}
  \end{minipage}
  \hfill
  \begin{minipage}[b]{0.45\textwidth}
    \includegraphics[width=\textwidth]{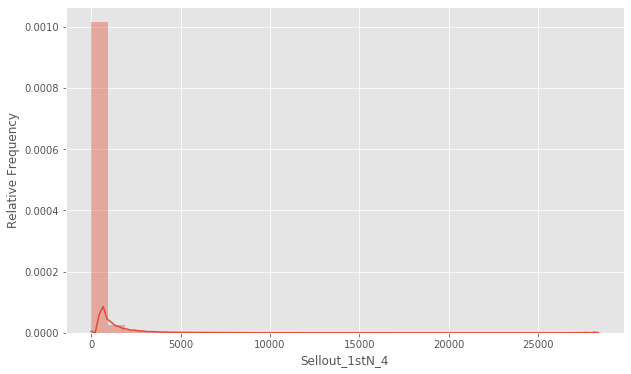}
    \caption{Distribution of Sell Out Neighbor Size-4 in 1.circle without zero values}
  \end{minipage}
\end{figure}

\vspace{6mm}
\begin{figure}[ht]
  \centering
  \begin{minipage}[b]{0.45\textwidth}
    \includegraphics[width=\textwidth]{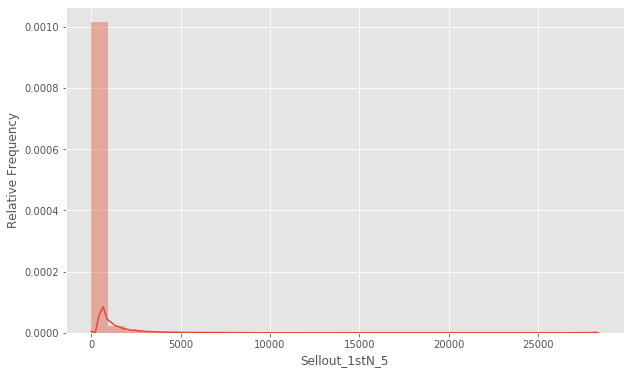}
    \caption{Distribution of Sell Out Neighbor Size-5 in 1.circle without zero values}
  \end{minipage}
  \hfill
  \begin{minipage}[b]{0.45\textwidth}
    \includegraphics[width=\textwidth]{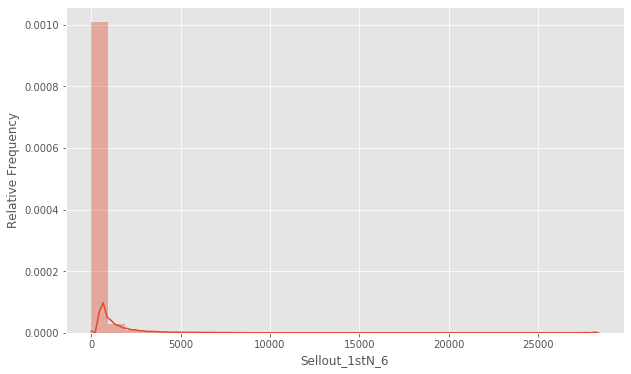}
    \caption{Distribution of Sell Out Neighbor Size-6 in 1.circle without zero values}
  \end{minipage}
\end{figure}

\vspace{6mm}
\begin{figure}[ht]
  \centering
  \begin{minipage}[b]{0.45\textwidth}
    \includegraphics[width=\textwidth]{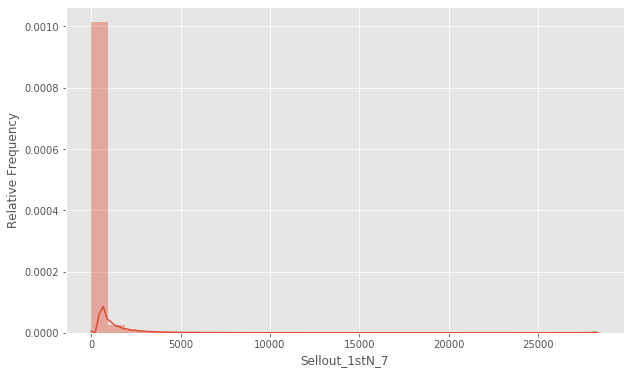}
    \caption{Distribution of Sell Out Neighbor Size-7 in 1.circle without zero values}
  \end{minipage}
  \hfill
  \begin{minipage}[b]{0.45\textwidth}
    \includegraphics[width=\textwidth]{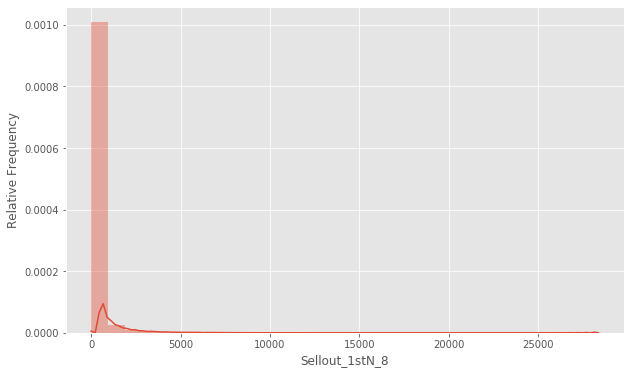}
    \caption{Distribution of Sell Out Neighbor Size-8 in 1.circle without zero values}
  \end{minipage}
\end{figure}

\vspace{6mm}
\begin{figure}[ht]
  \centering
  \begin{minipage}[b]{0.45\textwidth}
    \includegraphics[width=\textwidth]{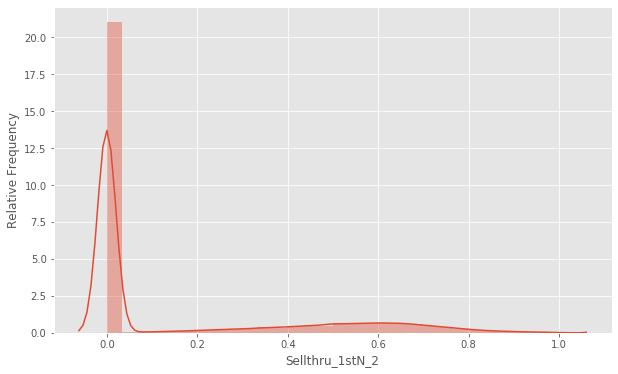}
    \caption{Distribution of Sell Through Neighbor Size-2 in 1.circle without zero values}
  \end{minipage}
  \hfill
  \begin{minipage}[b]{0.45\textwidth}
    \includegraphics[width=\textwidth]{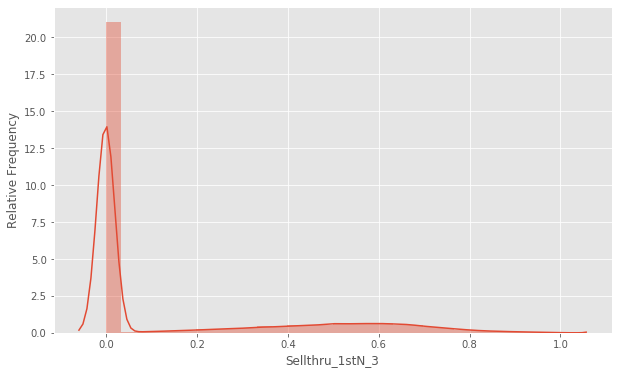}
    \caption{Distribution of Sell Through Neighbor Size-3 in 1.circle without zero values}
  \end{minipage}
\end{figure}

\vspace{6mm}
\begin{figure}[ht]
  \centering
  \begin{minipage}[b]{0.45\textwidth}
    \includegraphics[width=\textwidth]{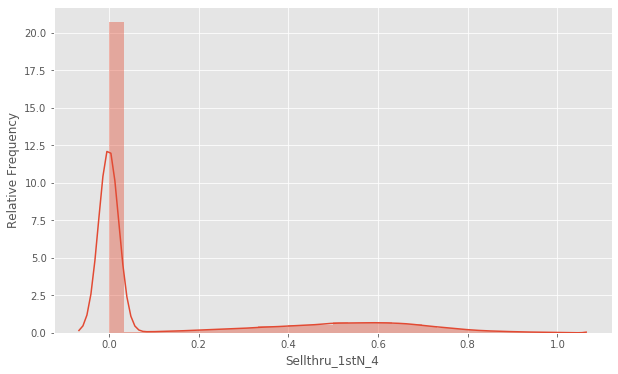}
    \caption{Distribution of Sell Through Neighbor Size-4 in 1.circle without zero values}
  \end{minipage}
  \hfill
  \begin{minipage}[b]{0.45\textwidth}
    \includegraphics[width=\textwidth]{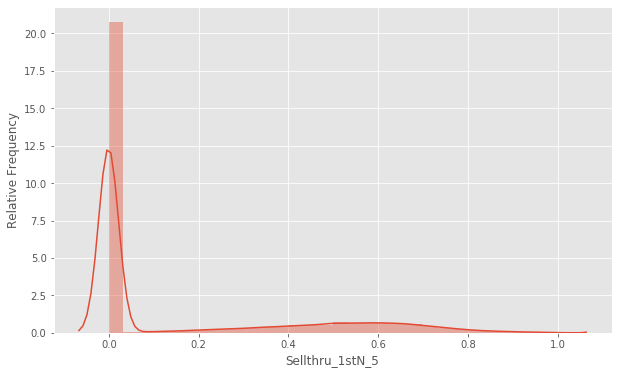}
    \caption{Distribution of Sell Through Neighbor Size-5 in 1.circle without zero values}
  \end{minipage}
\end{figure}

\vspace{6mm}
\begin{figure}[ht]
  \centering
  \begin{minipage}[b]{0.45\textwidth}
    \includegraphics[width=\textwidth]{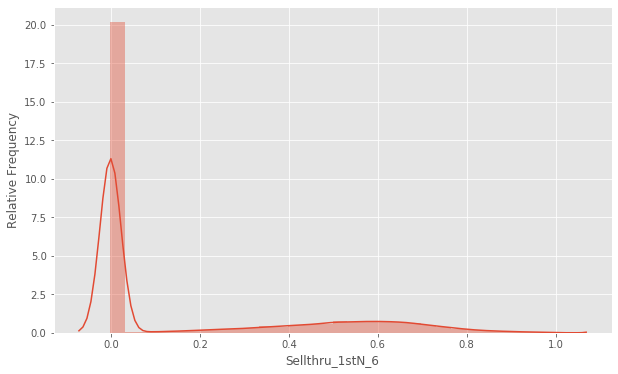}
    \caption{Distribution of Sell Through Neighbor Size-6 in 1.circle without zero values}
  \end{minipage}
  \hfill
  \begin{minipage}[b]{0.45\textwidth}
    \includegraphics[width=\textwidth]{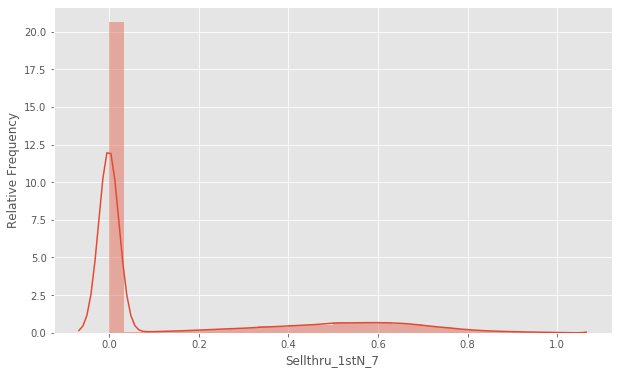}
    \caption{Distribution of Sell Through Neighbor Size-7 in 1.circle without zero values}
  \end{minipage}
\end{figure}

\vspace{6mm}
\begin{figure}[ht]
  \centering
  \begin{minipage}[b]{0.45\textwidth}
    \includegraphics[width=\textwidth]{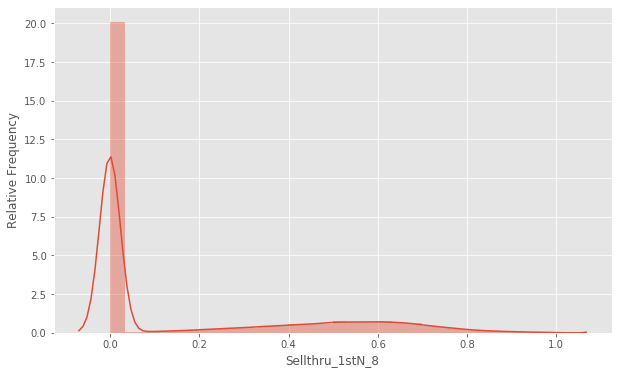}
    \caption{Distribution of Sell Through Neighbor Size-8 in 1.circle without zero values}
  \end{minipage}
  \hfill
\end{figure}

\vspace{6mm}
\begin{figure}[ht]
  \centering
  \begin{minipage}[b]{0.45\textwidth}
    \includegraphics[width=\textwidth]{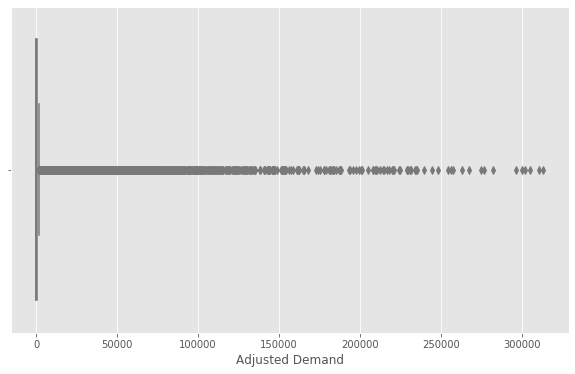}
    \caption{Box plot of Adjusted Demand without zero values}
  \end{minipage}
  \hfill
  \begin{minipage}[b]{0.45\textwidth}
    \includegraphics[width=\textwidth]{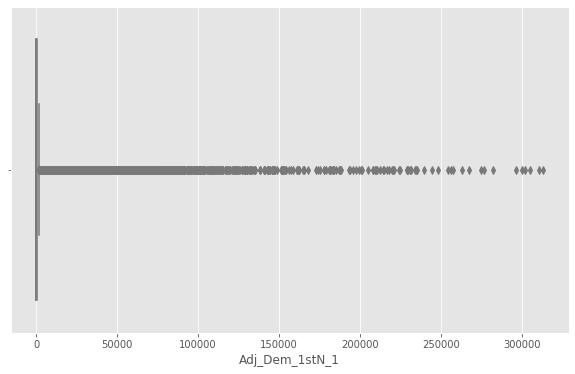}
    \caption{Box plot of Adjusted Demand Neighbor Size-1 in 1.circle without zero values}
  \end{minipage}
\end{figure}

\vspace{6mm}
\begin{figure}[ht]
  \centering
  \begin{minipage}[b]{0.45\textwidth}
    \includegraphics[width=\textwidth]{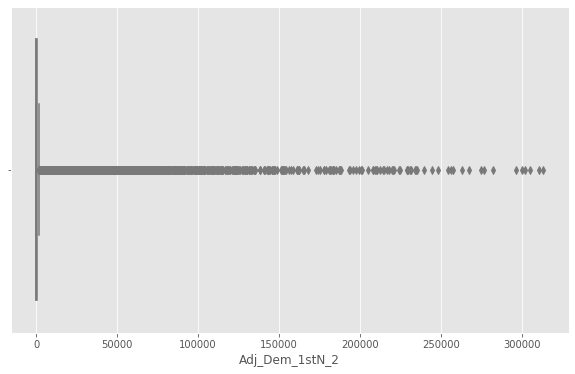}
    \caption{Box plot of Adjusted Demand Neighbor Size-2 in 1.circle without zero values}
  \end{minipage}
  \hfill
  \begin{minipage}[b]{0.45\textwidth}
    \includegraphics[width=\textwidth]{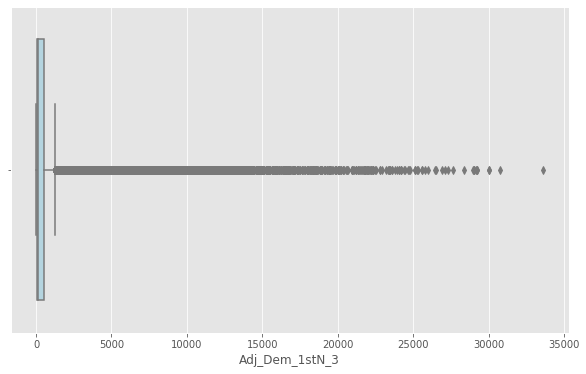}
    \caption{Box plot of Adjusted Demand Neighbor Size-3 in 1.circle without zero values}
  \end{minipage}
\end{figure}

\vspace{6mm}
\begin{figure}[ht]
  \centering
  \begin{minipage}[b]{0.45\textwidth}
    \includegraphics[width=\textwidth]{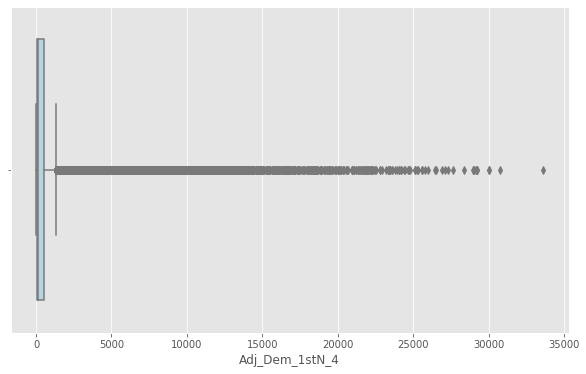}
    \caption{Box plot of Adjusted Demand Neighbor Size-4 in 1.circle without zero values}
  \end{minipage}
  \hfill
  \begin{minipage}[b]{0.45\textwidth}
    \includegraphics[width=\textwidth]{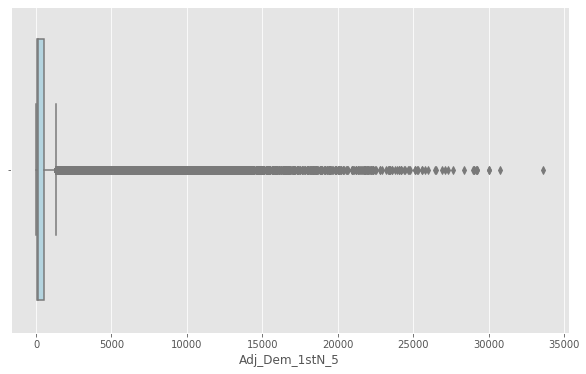}
    \caption{Box plot of Adjusted Demand Neighbor Size-5 in 1.circle without zero values}
  \end{minipage}
\end{figure}

\vspace{6mm}
\begin{figure}[ht]
  \centering
  \begin{minipage}[b]{0.45\textwidth}
    \includegraphics[width=\textwidth]{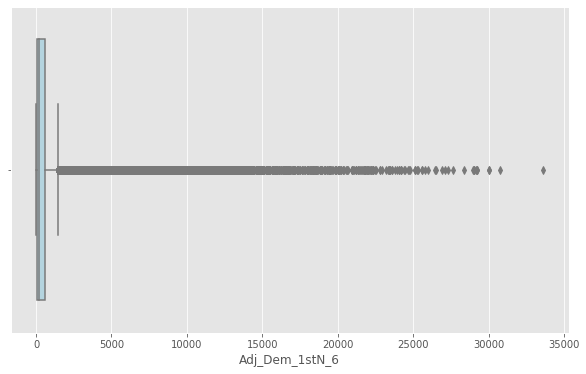}
    \caption{Box plot of Adjusted Demand Neighbor Size-6 in 1.circle without zero values}
  \end{minipage}
  \hfill
  \begin{minipage}[b]{0.45\textwidth}
    \includegraphics[width=\textwidth]{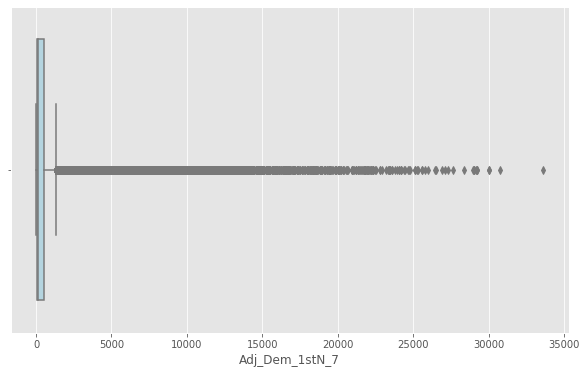}
    \caption{Box plot of Adjusted Demand Neighbor Size-7 in 1.circle without zero values}
  \end{minipage}
\end{figure}

\vspace{6mm}
\begin{figure}[ht]
  \centering
  \begin{minipage}[b]{0.45\textwidth}
    \includegraphics[width=\textwidth]{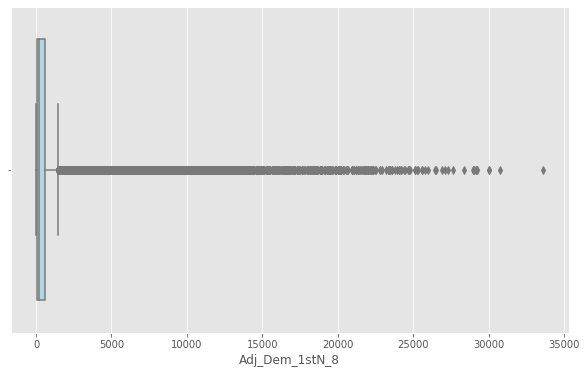}
    \caption{Box plot of Adjusted Demand Neighbor Size-8 in 1.circle without zero values}
  \end{minipage}
  \hfill
  \begin{minipage}[b]{0.45\textwidth}
    \includegraphics[width=\textwidth]{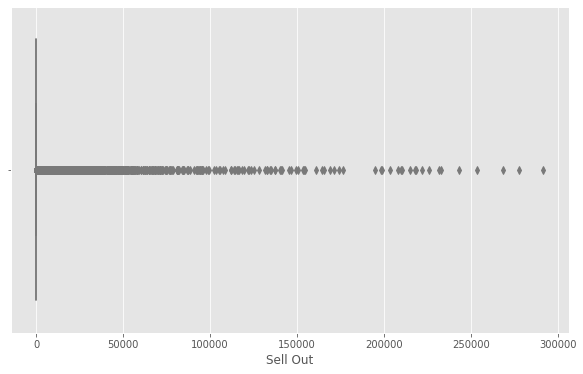}
    \caption{Box plot of Sell Out without zero values}
  \end{minipage}
\end{figure}

\vspace{6mm}
\begin{figure}[ht]
  \centering
  \begin{minipage}[b]{0.45\textwidth}
    \includegraphics[width=\textwidth]{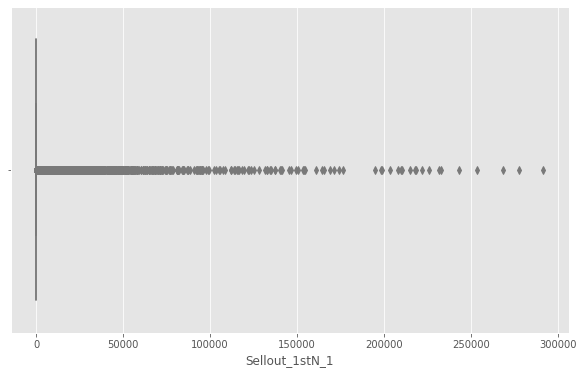}
    \caption{Box plot of Sell Out Neighbor Size-1 in 1.circle without zero values}
  \end{minipage}
  \hfill
  \begin{minipage}[b]{0.45\textwidth}
    \includegraphics[width=\textwidth]{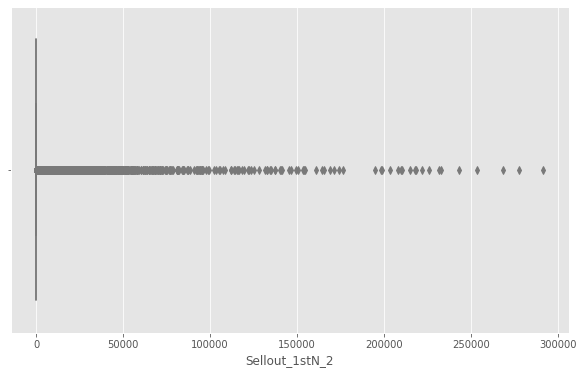}
    \caption{Box plot of Sell Out Neighbor Size-2 in 1.circle without zero values}
  \end{minipage}
\end{figure}

\vspace{6mm}
\begin{figure}[ht]
  \centering
  \begin{minipage}[b]{0.45\textwidth}
    \includegraphics[width=\textwidth]{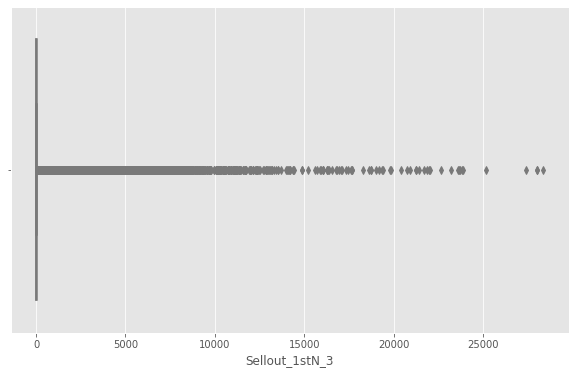}
    \caption{Box plot of Sell Out Neighbor Size-3 in 1.circle without zero values}
  \end{minipage}
  \hfill
  \begin{minipage}[b]{0.45\textwidth}
    \includegraphics[width=\textwidth]{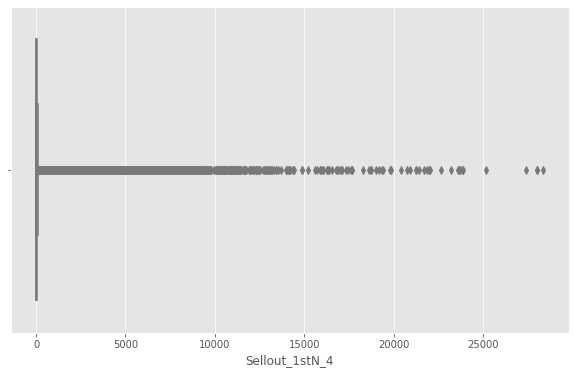}
    \caption{Box plot of Sell Out Neighbor Size-4 in 1.circle without zero values}
  \end{minipage}
\end{figure}

\vspace{6mm}
\begin{figure}[ht]
  \centering
  \begin{minipage}[b]{0.45\textwidth}
    \includegraphics[width=\textwidth]{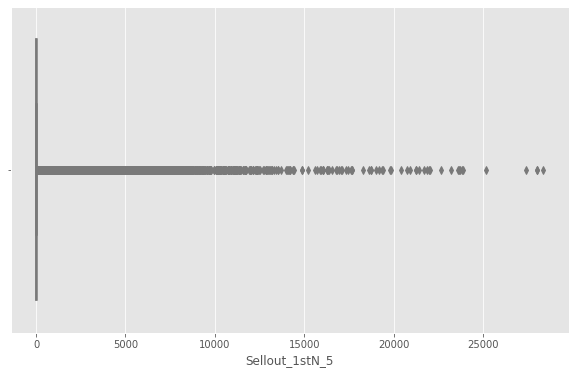}
    \caption{Box plot of Sell Out Neighbor Size-5 in 1.circle without zero values}
  \end{minipage}
  \hfill
  \begin{minipage}[b]{0.45\textwidth}
    \includegraphics[width=\textwidth]{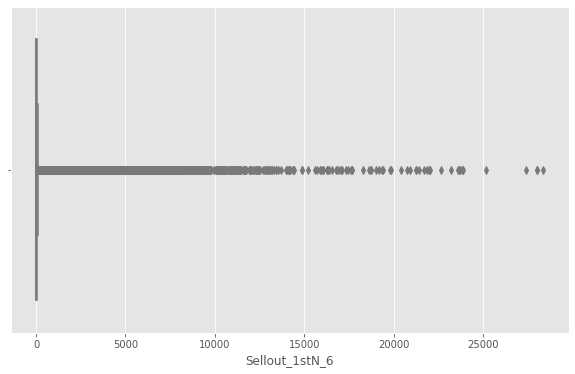}
    \caption{Box plot of Sell Out Neighbor Size-6 in 1.circle without zero values}
  \end{minipage}
\end{figure}

\vspace{6mm}
\begin{figure}[ht]
  \centering
  \begin{minipage}[b]{0.45\textwidth}
    \includegraphics[width=\textwidth]{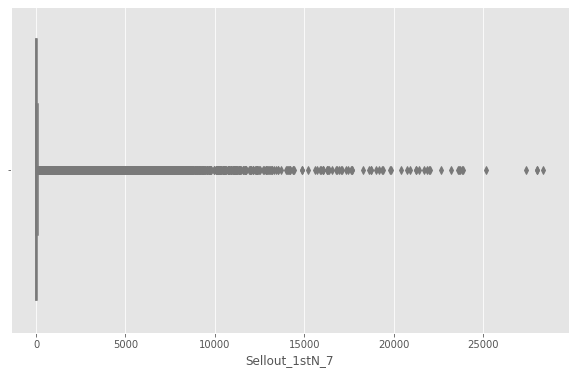}
    \caption{Box plot of Sell Out Neighbor Size-7 in 1.circle without zero values}
  \end{minipage}
  \hfill
  \begin{minipage}[b]{0.45\textwidth}
    \includegraphics[width=\textwidth]{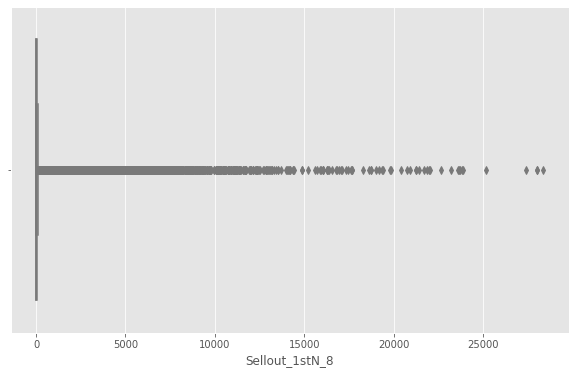}
    \caption{Box plot of Sell Out Neighbor Size-8 in 1.circle without zero values}
  \end{minipage}
\end{figure}

\vspace{6mm}
\begin{figure}[ht]
  \centering
  \begin{minipage}[b]{0.45\textwidth}
    \includegraphics[width=\textwidth]{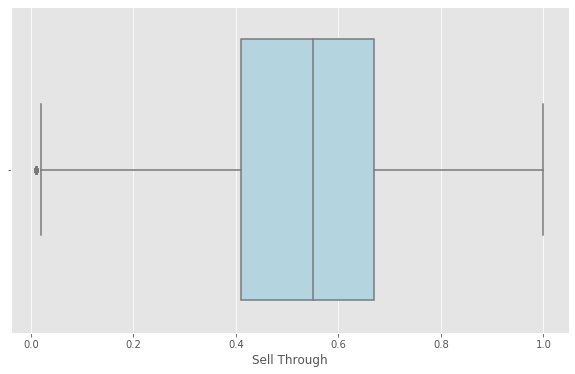}
    \caption{Box plot of Sell through without zero values}
  \end{minipage}
  \hfill
  \begin{minipage}[b]{0.45\textwidth}
    \includegraphics[width=\textwidth]{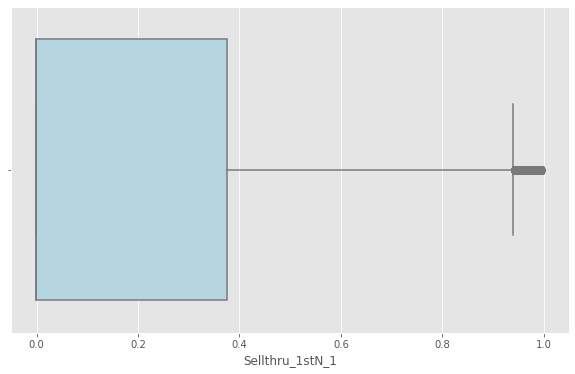}
    \caption{Box plot of Sell through Neighbor Size-1 in 1.circle without zero values}
  \end{minipage}
\end{figure}

\vspace{6mm}
\begin{figure}[ht]
  \centering
  \begin{minipage}[b]{0.45\textwidth}
    \includegraphics[width=\textwidth]{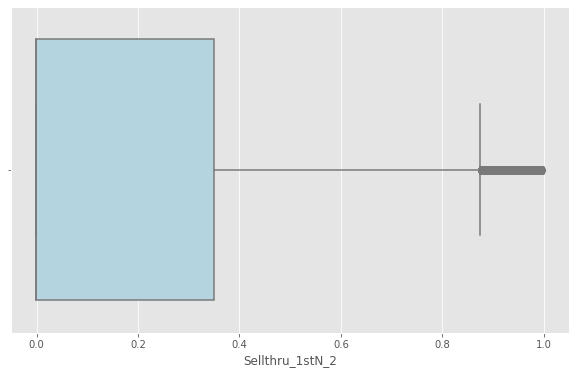}
    \caption{Box plot of Sell through Neighbor Size-2 in 1.circle without zero values}
  \end{minipage}
  \hfill
  \begin{minipage}[b]{0.45\textwidth}
    \includegraphics[width=\textwidth]{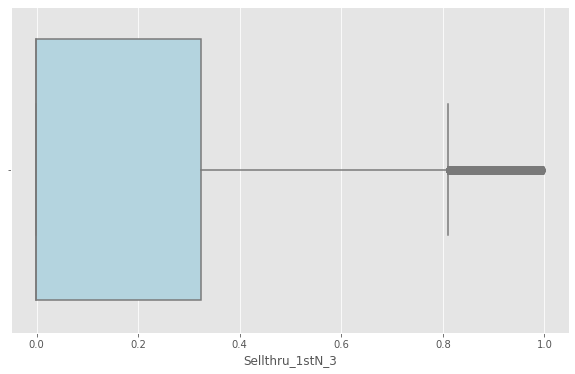}
    \caption{Box plot of Sell through Neighbor Size-3 in 1.circle without zero values}
  \end{minipage}
\end{figure}

\vspace{6mm}
\begin{figure}[ht]
  \centering
  \begin{minipage}[b]{0.45\textwidth}
    \includegraphics[width=\textwidth]{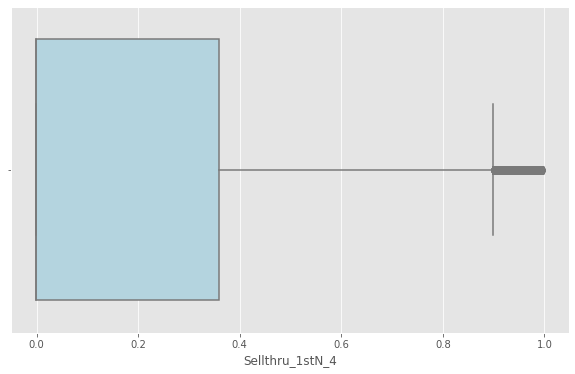}
    \caption{Box plot of Sell through Neighbor Size-4 in 1.circle without zero values}
  \end{minipage}
  \hfill
  \begin{minipage}[b]{0.45\textwidth}
    \includegraphics[width=\textwidth]{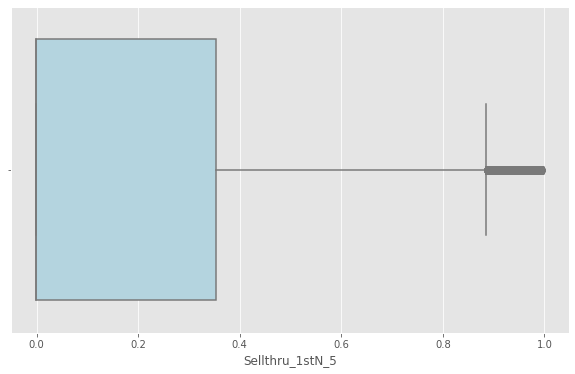}
    \caption{Box plot of Sell through Neighbor Size-5 in 1.circle without zero values}
  \end{minipage}
\end{figure}

\vspace{6mm}
\begin{figure}[ht]
  \centering
  \begin{minipage}[b]{0.45\textwidth}
    \includegraphics[width=\textwidth]{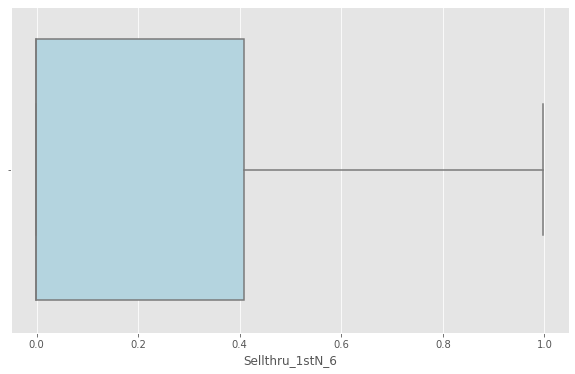}
    \caption{Box plot of Sell through Neighbor Size-6 in 1.circle without zero values}
  \end{minipage}
  \hfill
  \begin{minipage}[b]{0.45\textwidth}
    \includegraphics[width=\textwidth]{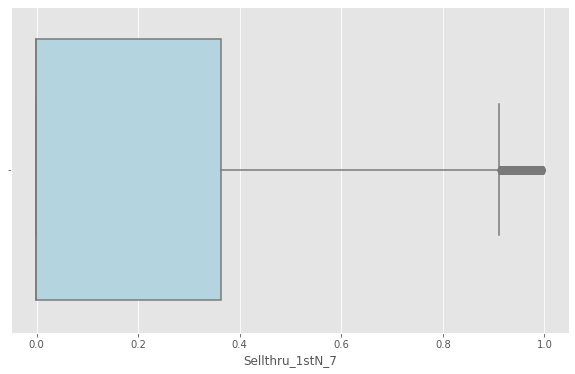}
    \caption{Box plot of Sell through Neighbor Size-7 in 1.circle without zero values}
  \end{minipage}
\end{figure}

\vspace{6mm}
\begin{figure}[ht]
  \centering
  \begin{minipage}[b]{0.45\textwidth}
    \includegraphics[width=\textwidth]{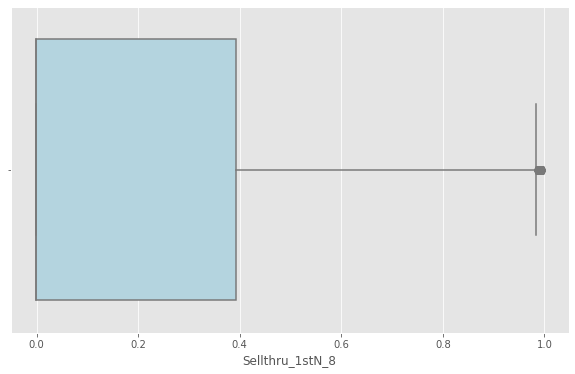}
    \caption{Box plot of Sell through Neighbor Size-8 in 1.circle without zero values}
  \end{minipage}
  \hfill
\end{figure}

\vspace{2mm}
\begin{figure}[ht]
  \centering
  \begin{minipage}[b]{1.1\textwidth}
    \includegraphics[width=\textwidth]{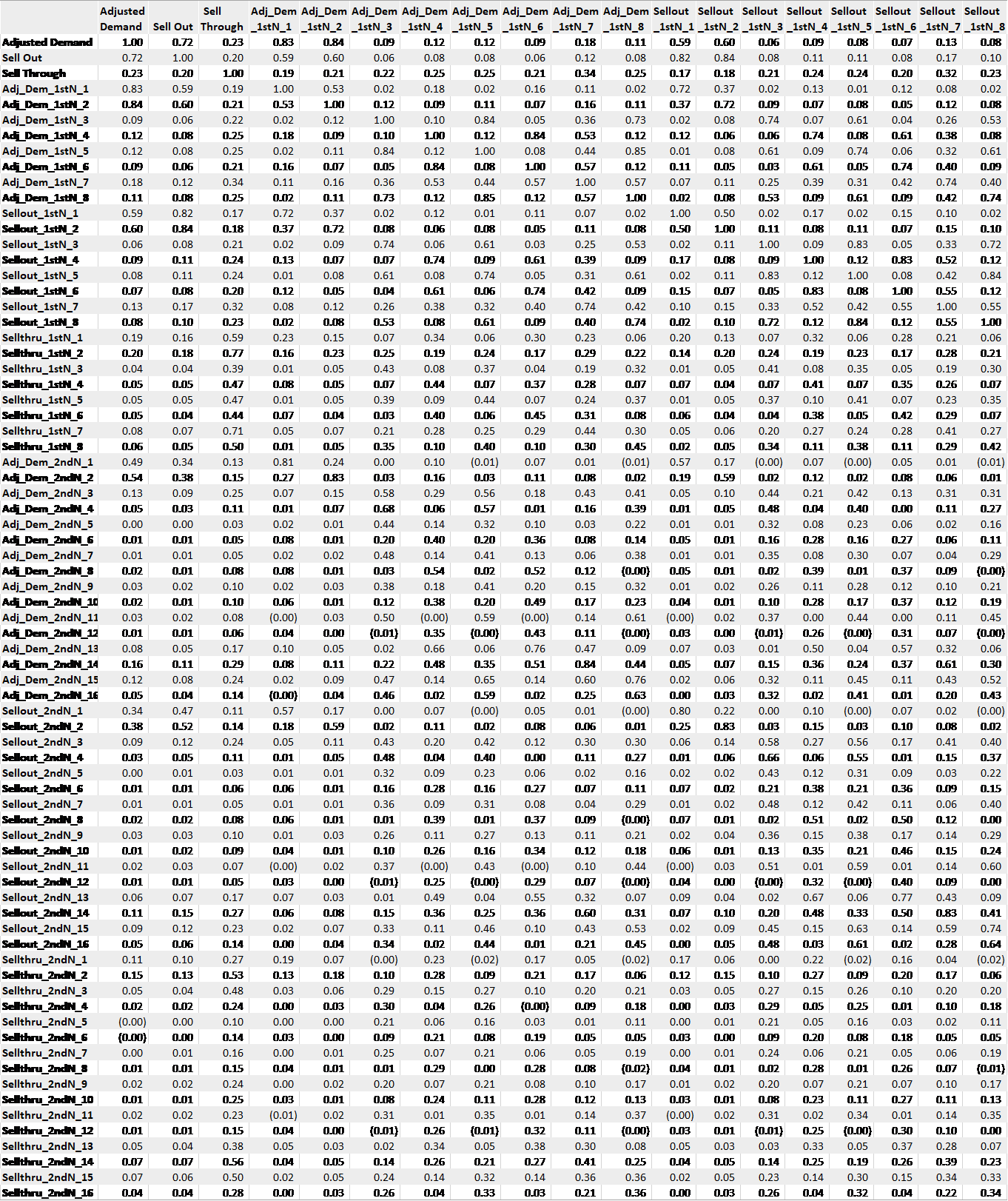}
    \caption{Correlation Matrix I}
  \end{minipage}
\end{figure}

\vspace{2mm}
\begin{figure}[ht]
  \centering
  \begin{minipage}[b]{1.1\textwidth}
    \includegraphics[width=\textwidth]{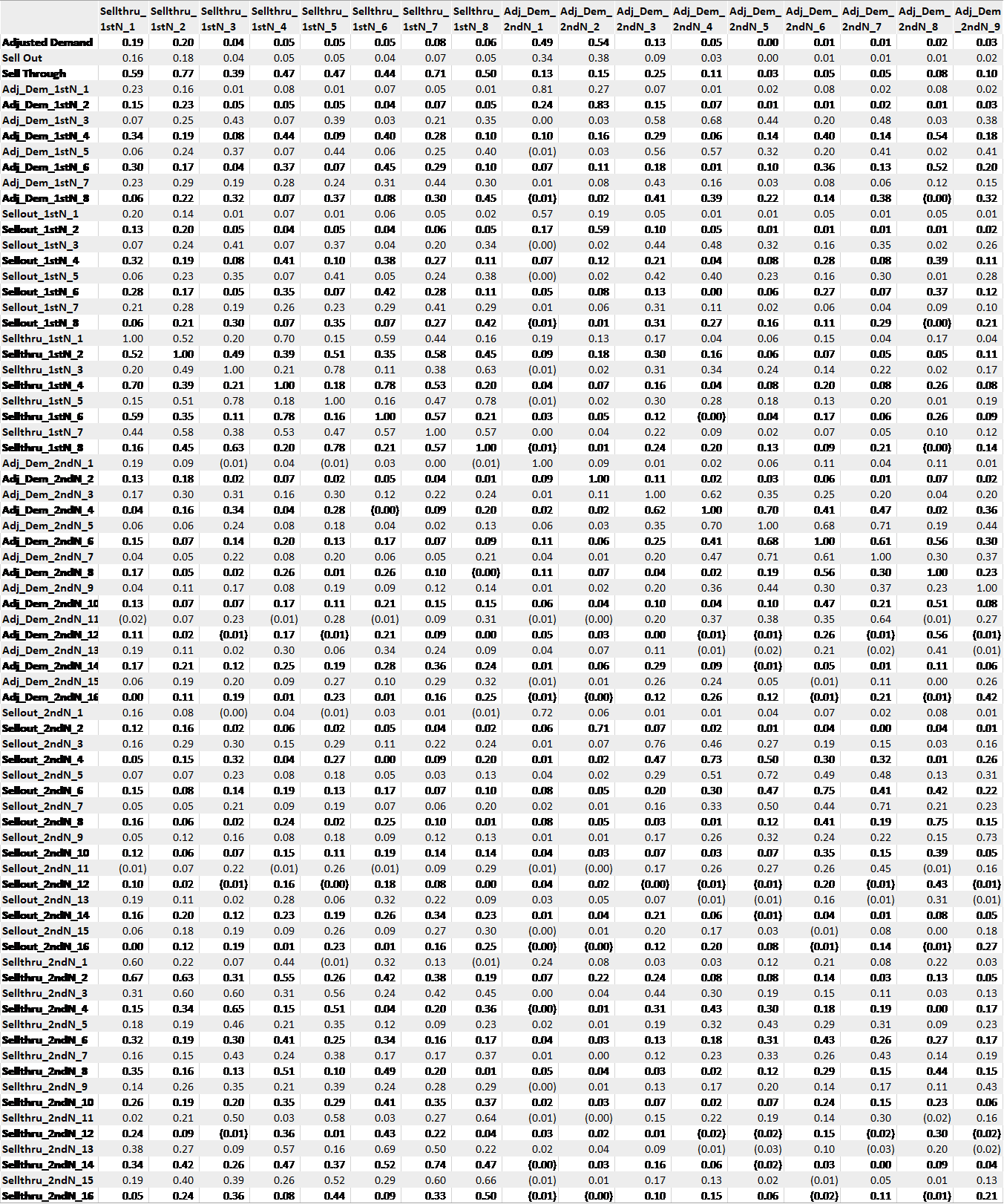}
    \caption{Correlation Matrix II}
  \end{minipage}
\end{figure}

\vspace{2mm}
\begin{figure}[ht]
  \centering
  \begin{minipage}[b]{1.1\textwidth}
    \includegraphics[width=\textwidth]{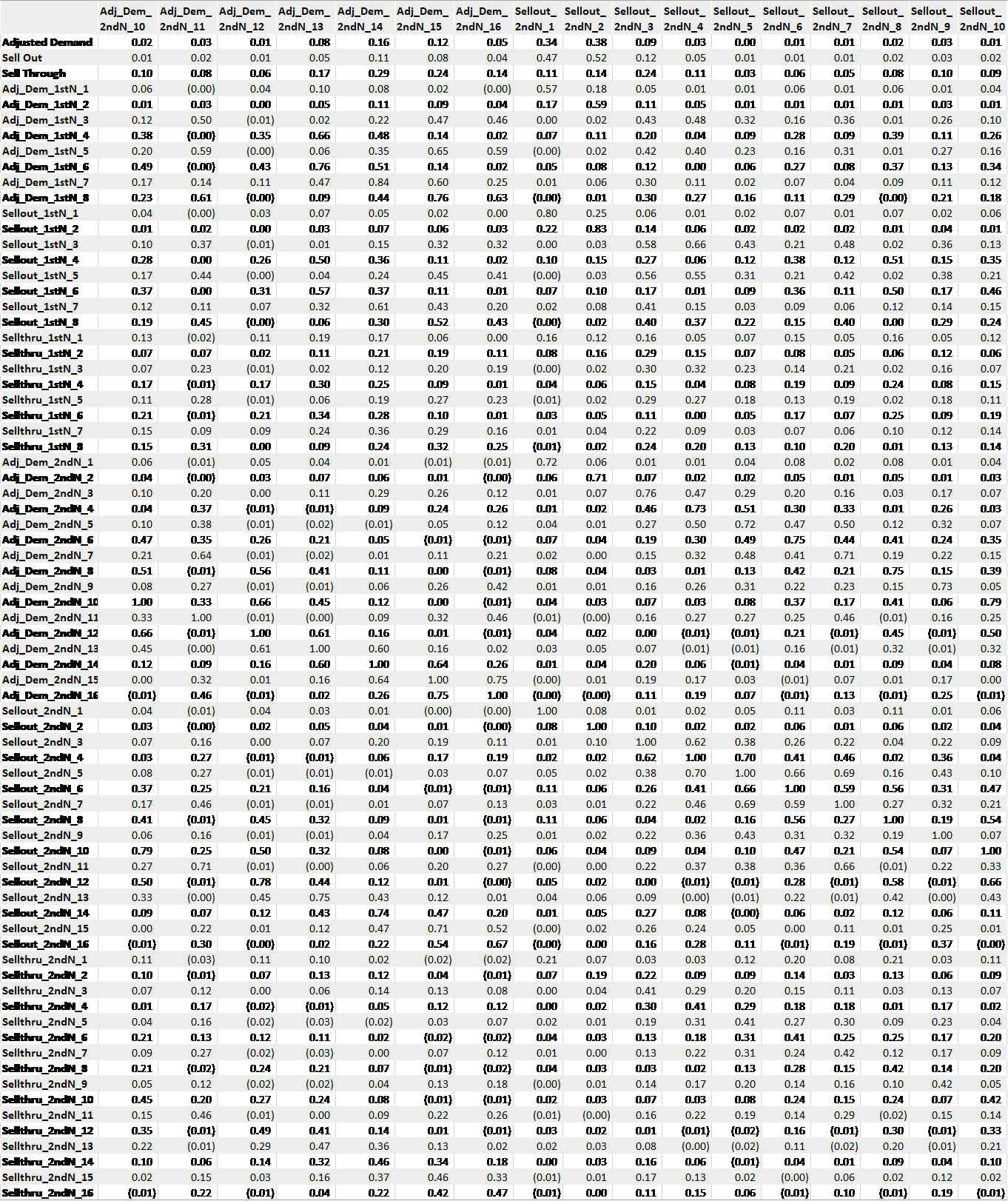}
    \caption{Correlation Matrix III}
  \end{minipage}
\end{figure}

\vspace{2mm}
\begin{figure}[ht]
  \centering
  \begin{minipage}[b]{1\textwidth}
    \includegraphics[width=\textwidth]{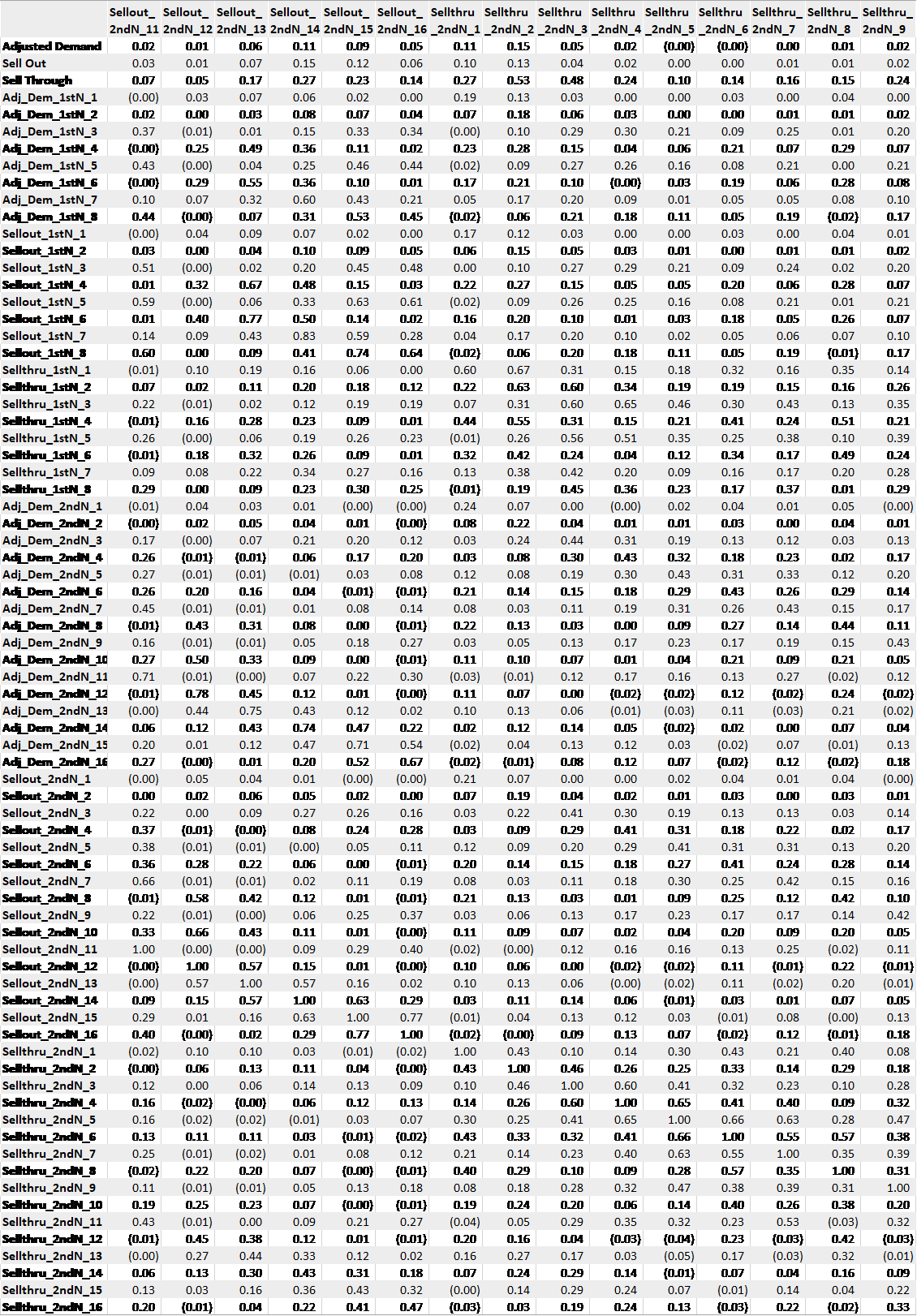}
    \caption{Correlation Matrix IV}
  \end{minipage}
\end{figure}

\vspace{2mm}
\begin{figure}[ht]
  \centering
  \begin{minipage}[b]{0.5\textwidth}
    \includegraphics[width=\textwidth]{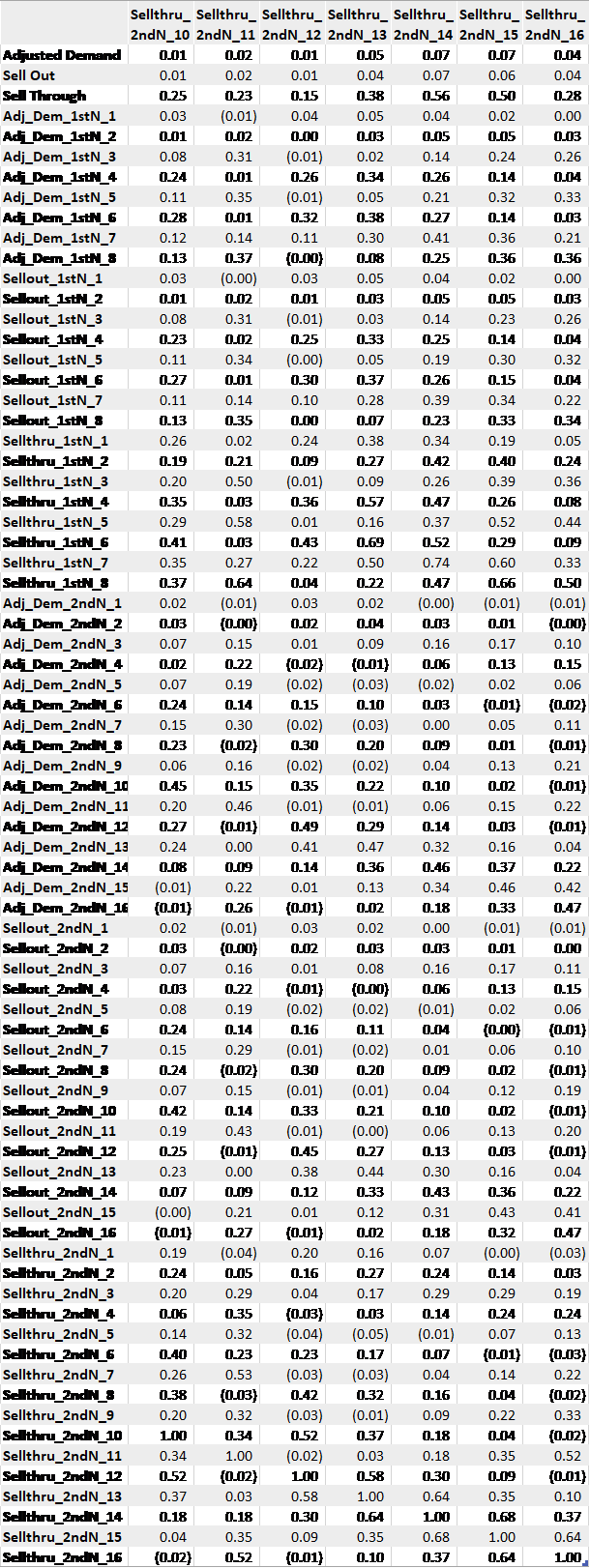}
    \caption{Correlation Matrix V}
  \end{minipage}
\end{figure}

\end{document}